\documentclass[3p,authoryear, final]{elsarticle}
\usepackage{amssymb}
\usepackage{amsthm}
\usepackage{amsmath}

\usepackage{lineno}
\usepackage[round]{natbib}
\usepackage{tabularx}
\usepackage{array}
\usepackage{tikz}
\usetikzlibrary{shapes,arrows,positioning,calc,backgrounds}
\usepackage{ifthen}
\usepackage[]{algorithm2e}
\usepackage[section]{placeins}
\usepackage{graphicx}
\usepackage{caption}
\usepackage{standalone}
\usepackage{float}
\floatstyle{plaintop}
\restylefloat{table}
\usepackage{subcaption}
\usepackage{booktabs}
\usepackage{wrapfig}
\usepackage{url}
\usepackage{lscape}
\usepackage{soul}
\usepackage{color}

\soulregister\cite7
\soulregister\citet7
\soulregister\citep7
\soulregister\ref7
\usepackage{setspace}
\usepackage[tableposition=top]{caption}
\usepackage{multirow}
\journal{International Journal of Applied Earth Observation and Geoinformation}
\definecolor{green}{RGB}{0,168,89}
\definecolor{yellow}{RGB}{255,204,41}
\definecolor{blue}{RGB}{62,64,149}

\begin{document}
\doublespacing
\tikzset{
    position/.style args={#1:#2 from #3}{
        at=(#3.#1), anchor=#1+180, shift=(#1:#2)
    }
}	
	\tikzstyle{blockfor} = [circle, draw, fill=green!40, 
	text width=2em, text centered, minimum height=2em]
	\tikzstyle{blockdef} = [circle, draw, fill=red!40, 
	text width=2em, text centered, minimum height=2em]
	\tikzstyle{blockpas} = [circle, draw, fill=yellow!40, 
	text width=2em, text centered, minimum height=2em]
	\tikzstyle{line} = [draw, -latex']	
	\tikzstyle{state} = [draw, very thick, fill=white, rectangle, minimum height=3em, minimum width=7em, node distance=8em, font={\sffamily\bfseries}]
	\tikzstyle{stateEdgePortion} = [black,thick]
	\tikzstyle{stateEdge} = [stateEdgePortion,->]
	\tikzstyle{stateEdgeBoth} = [stateEdgePortion,<->]

\newcommand{\marker}[3]{
  \tikz[baseline=(X.base)]{
    \node [fill=#1!40,rounded corners] (X) {#2:};
  }
  {\color{#1!80!black}#3}
}
\newcommand{\todo}[1]{\marker{red}{Todo}{#1}}
\renewcommand{\check}[1]{\marker{cyan}{Check}{#1}}
\newcommand{\ribana}[1]{\marker{magenta}{Ribana}{#1}}

\newcommand{\etal}{et~al.}
\newcommand{\wrt}{w.r.t.}
\newcommand{\ie}{i.e.}
\newcommand{\Ie}{I.e.}
\newcommand{\eg}{e.g.}
\newcommand{\Eg}{E.g.}
\newcommand{\cf}{c.f.}

\renewcommand{\(}{\left(}
\renewcommand{\)}{\right)}
\renewcommand{\d}[1]{{\mbox{\boldmath$#1$}}}
\newcommand{\m}[1]{{\mbox{{\fontencoding{T1}\sffamily\slshape{#1\/}}}}}
\newcommand{\trans}[0]{^{\sf T}}
\newcommand{\ical}[1]{{\mbox{\usefont{OT1}{pzc}{m}{it}{#1}}}}
\newcommand{\Diag}[0]{{\mbox{Diag}}}
\newcommand{\inv}[0]{^{-1}}

\begin{frontmatter}
\title{Tropical Land Use Land Cover Mapping in Par\'{a} (Brazil) using Discriminative Markov Random Fields and Multi-temporal TerraSAR-X Data}
\author[ber]{Ron Hagensieker\corref{cor}}\ead{ron.hagensieker@fu-berlin.de}
\author[bon]{Ribana Roscher}\ead{ribana.roscher@uni-bonn.de}
\author[ber]{Johannes Rosentreter}\ead{johannes.rosentreter@fu-berlin.de}
\author[hub]{Benjamin Jakimow}\ead{benjamin.jakimow@geo.hu-berlin.de}
\author[ber]{Bj\"orn Waske}\ead{bjoern.waske@fu-berlin.de}
\address[ber]{Freie Universit\"at Berlin, Institute of Geographical Sciences, Malteserstr. 74-100, 12249 Berlin, Germany}
\address[bon]{Rheinische Friedrich-Wilhelms-Universit\"at Bonn, Institute of Geodesy and Geoinformation, Nussallee 15, 53115 Bonn, Germany}
\address[hub]{Humboldt-Universit\"at zu Berlin, Geography Department, Unter den Linden 6, 10099 Berlin, Germany}
\cortext[cor]{Corresponding author. Tel.: (+49 30) 838- 458 683}
\begin{abstract}
Remote sensing satellite data offer the unique possibility to map land use land cover transformations by providing spatially explicit information. However, detection of short-term processes and land use patterns of high spatial-temporal variability is a challenging task.\\We present a novel framework using multi-temporal TerraSAR-X data and machine learning techniques, namely Discriminative Markov Random Fields with spatio-temporal priors, and Import Vector Machines, in order to advance the mapping of land cover characterized by short-term changes. Our study region covers a current deforestation frontier in the Brazilian state Par\'{a} with land cover dominated by primary forests, different types of pasture land and secondary vegetation, and land use dominated by short-term processes such as slash-and-burn activities. The data set comprises multi-temporal TerraSAR-X imagery acquired over the course of the 2014 dry season, as well as optical data (RapidEye, Landsat) for reference. Results show that land use land cover is reliably mapped, resulting in spatially adjusted overall accuracies of up to $79\%$ in a five class setting, yet limitations for the differentiation of different pasture types remain.\\The proposed method is applicable on multi-temporal data sets, and constitutes a feasible approach to map land use land cover in regions that are affected by high-frequent temporal changes.
\end{abstract}
\begin{keyword}
Markov Random Fields (MRF) \sep Import Vector Machines (IVM) \sep Multi-temporal LULC mapping \sep Deforestation \sep Amazon \sep SAR
\end{keyword}
\end{frontmatter}
\section{Introduction}
The Brazilian Amazon is the largest area of tropical rain forest shared by a single country. 
In the last decades it has become increasingly threatened by large scale deforestation, forest degradation, and the expansion of agriculture \citep{davidson2012amazon,lapola2014pervasive}.
They affect the Earth's ecosystems and ecosystem services far beyond the boundaries of the original region, and can influence the climate directly at local and even regional scales \citep{Foley2005,Vitousek1997}.
Thus, detailed knowledge and information on land use and land cover (LULC) offers valuable input for decision support and environmental monitoring systems.
\\Remote sensing satellite data offers the unique possibility to generate consistent LULC maps over large areas at a temporally high resolution.
Mapping of LULC change in the Amazon is predominantly achieved by analyzing multi-spectral remote sensing data \citep{inpe2015projetoprodes,Wulder2012opening,Hansen850highresolution}.
However, a limitation of the analysis of multi-spectral remote sensing data is imposed by its dependency on cloud-free conditions.
These are rare in tropical regions and in general not met during wet season \citep[e.g.][]{Rufin2015,Mueller2015}.
Synthetic aperture radar (SAR) data can overcome these problems and various studies demonstrate the potential for mapping LULC and their changes \citep{Pfeifer2016,Qi2012,Bovolo2005}, also in the context of deforestation and related processes \citep{sarker2013forest,reiche2015fusing,Englhart2011,Almeida?Filho2009}.
Such mapping approaches become even more attractive due to recent missions with increased repetition rates, higher spatial resolution (e.g. TerraSAR-X and Sentinel-1), as well as better data availability, e.g., by virtue of the Copernicus data policy \citep{Aschbacher2012}. 
TerraSAR-X and the Sentinel-1 constellation guarantee cloud free coverage within 11 and 6 days respectively, while the repetition rate of the Sentinel-2 constellation (5 days) and Landsat-8 (16 days) might be affected by clouds.
\\Although the classification accuracy of SAR data can be limited in direct comparison to multi-spectral data, various approaches exist to increase the mapping accuracy. 
These include the integration of one-pass interferometry \citep{schlund2013importance}, contextual spatial information derived from texture parameters or segmentation \citep{cutler2012estimating,sarker2013forest,schlund2013importance,waske2008classifying}, or the utilization of multi-temporal or multi-sensoral data \citep{Reiche2013,stefanski2014mapping,waske2009classifier}. 
Although limitations of short wavelength SAR data for the classification of dense vegetation are well documented \citep[e.g.][]{kumar2013discrimination}, various studies have highlighted the potentials of this data for LULC mapping \citep[e.g.][]{schlund2013importance,Qi2015,Uhlmann2014,Qi2012,Khatami2016,Sonobe2014}, e.g. by utilization of multi-temporal data, modern classification algorithms, or spatial context.
Multi-temporal data sets are generally more adequate when classes can be characterized by clearly defined temporal signatures, e.g. caused by differences in the phenology of crops, land use management, or seasonal cycles \citep{blaes2005efficiency,mcnairn2009integration}.
While the single classification of a multitemporal data set might be useful for study sites without or long-term changes \citep{waske2009classifier,Stefanski2014}, it might be limited for study sites with temporally high-frequent changes in land cover, e.g. slash-and-burn activities, at arbitrary points in time.
Recent studies have shown great potentials to tackle these problems by time series analysis of multispectral data \citep{zhu2014continuous}, but SAR speckle and quick succession processes still pose difficult challenges using such methods, especially if very long time series are often not available.\\
In the context of multi-temporal data analysis, a main drawback is often the assumption of non changing land cover during the investigation period. Consequently, temporally dynamic LULC, such as slash-and-burn activities or transitions between clean and shrubby pasture, are neglected.
Various studies emphasize the usage of an adequate classification approach to ensure a high mapping accuracy \citep{liu2006spatial,waske2007fusion,waske2009classifier}. Especially the integration of spatial information by means of region-based classification or spatial features such as texture lead to a gain in accuracy. In addition, Markov Random Fields (MRFs) are a promising approach to integrate spatial context \citep{moser2013land,moser2013combining,liu2006spatial}. MRFs are employed to model prior knowledge about neighborhood relations within the image, called spatial relations, but can also be extended to describe relations of the same area but at different acquisition dates (temporal relations).
Since the early 1990s, approaches based on MRFs have been utilized in remote sensing for various purposes \citep{bouman1994multiscale,xie2002sar,tran2005init,solberg1996markov}.
\citet{liu2008using} use locally variant transition models to account for spatial heterogeneity and have applied the model on subsets of two Landsat scenes from 1990 and 2001. More recently, \citet{wehmann2015spatial} have adapted an integrated kernel as proposed by \citet{moser2013combining}, and used Iterative Conditional Modes (ICM) as optimization technique with spatially-variant transitions for classifying Landsat data. 
\citet{hoberg2015conditional} apply multi-temporal Conditional Random Fields to regularize annual remote sensing imagery from different high resolution scales (IKONOS, RapidEye, Landsat) over the course of five years.
\\With the emergence of efficient probabilistic classifiers over the last decade, standard MRFs have been extended to discriminative MRFs \citep{Kumar2003}, and turn out to be increasingly useful to optimize land cover classifications \citep{moser2010contextual,tarabalka2010svm,Voisin2013}.
\citet{liu2006spatial} highlight the advantages of utilizing non-parametric, probabilistic Support Vector Machines (SVMs, \citet{platt1999probabilistic}) over a maximum likelihood classifier.
However, although many remote sensing studies highlight the positive capabilities of MRFs, only few studies aim on using MRFs for landscape-scale mapping with multi-temporal data sets \citep[e.g.][]{Cai2014,wehmann2015spatial,Olding2015}, for example, to map forest cover change \citep{liu2008using,liu2006spatial}.
If multi-temporal data sets are available, MRFs can also be used to optimize the corresponding maps by considering predefined spatial-temporal inter-dependencies between neighboring pixels, which are stored in transition matrices.
\\We present a novel framework for classification of a TS-X time series using discriminative MRFs and Import Vector Machine (IVM), a probabilistic, discriminative, non-parametric classifier. Each scene is separately classified using IVM, afterwards MRFs are utilized in an independent step to post-regularize the classification map. We chose IVMs over commonly used probabilistic SVMs, since they have proven to offer a more reliable probabilistic output \citep{zhu2005kernel,roscher2012sup,roscher2012incremental}. For MRF optimization we choose Loopy Belief Propagation (LBP) over ICM as this method has been shown repeatedly to yield higher accuracies \citep{Szeliski2006,Andres2010}. Few studies have utilized LBP in the field of remote sensing \citep{Li2013}, and as a novelty we integrate LBP into a multi-temporal setting.
\\The presented framework aims on the classification of each individual acquisition, and thus enables mapping of high frequency spatial-temporal LULC patterns. In contrast to related studies, we use a multi-temporal MRF model on SAR data to detect short-term transitions within one season and Loopy Belief Propagation (LBP) for inference.
\\The overall goal of this research is focused on two objectives: (i) to map LULC in a tropical setting with short-term processes, by adapting recent MRF methods, and (ii) to assess the potential for LULC mapping using time-series image data of short wavelength SAR.
The specific objective is to map LULC in Par\'{a}, Brazil, where transformations of forest to pasture are the major driver of deforestation. Pasture management in the study region tends to fall into one of two categories: long-term processes of intensively managed pasture land (pasto limpo), or short-term processes of episodically managed pasture land with a high degree of successive dynamics (pasto sujo). Pasture management in general is characterized by slash-and-burn processes resulting in sudden changes in LULC.
\section{Study Area \& Data}
\subsection{Study Area}
\label{datasa}
\begin{figure*}
   \centering
	\includegraphics[width=0.8301\textwidth]{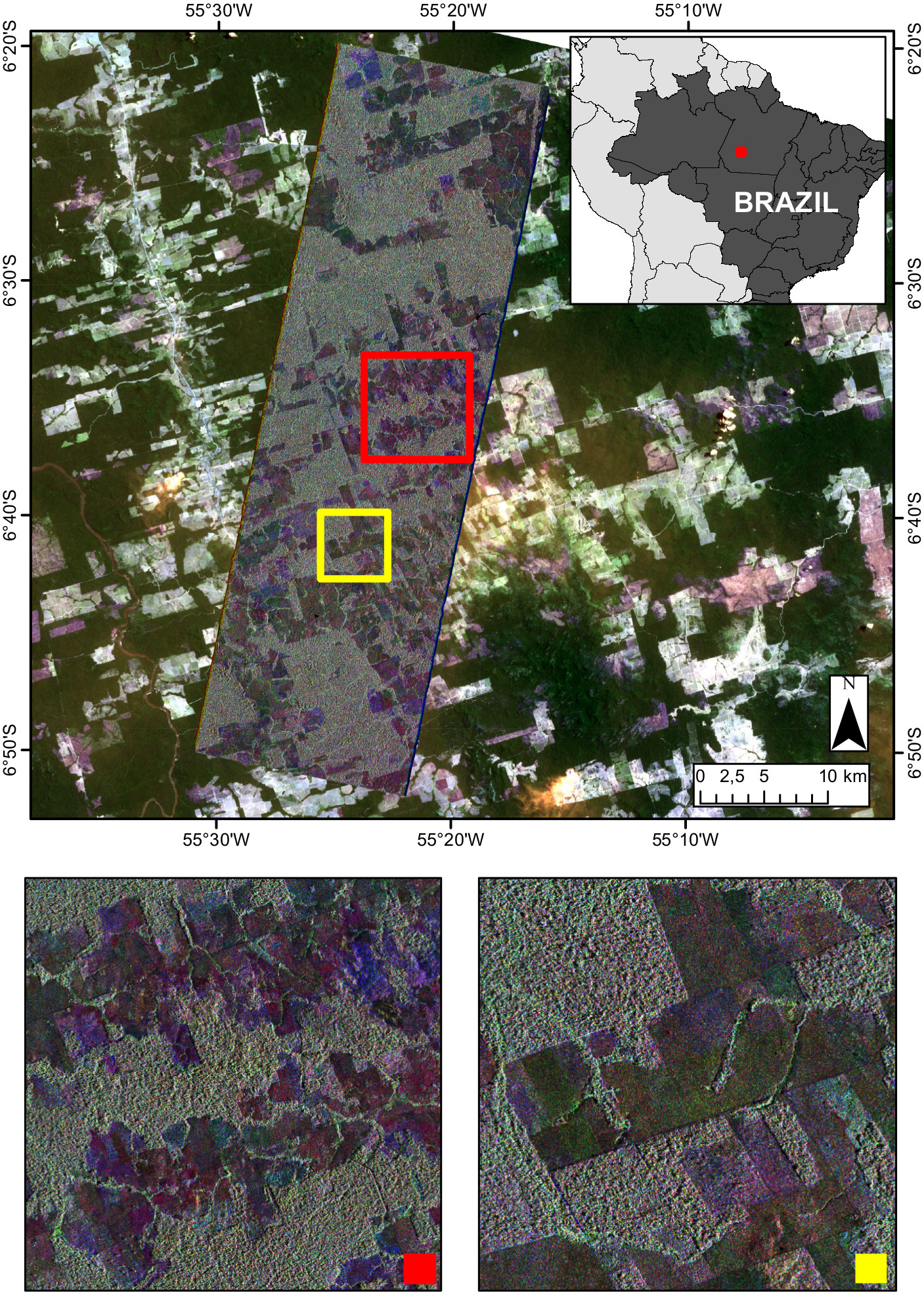}
	\caption{\label{fig:tsxsaarea}Composite of three TerraSAR-X acquisitions (Red: VV June 8, Green: HH September 4, Blue: VV November 9 of 2014). True color ETM (L8, 12 September 2014) in the background shows the diverse LULC properties.}
\end{figure*}
The study area lies in the Northern part of the Novo Progresso municipality (southern Par\'{a} state, Brazil), and is intersected by the BR-163 highway in the Southwest \ref{fig:tsxsaarea}.
The BR-163 is accompanied by fishbone structures indicative of deforestation \citep{Ahmed2013,coy2014frentes}.
A major driver of deforestation in the study area is the transformation of forests into pasture land.
The climate in the study region is characterized by a wet and a dry season. While the dry season, between June and September, sees abrupt land cover changes in the form of large scale burning and clear cuts, the wet season is defined by gradual regrowth, yet deforestation rates over the wet season are on the rise.
\begin{figure}[!ht]
\centering
\includegraphics[width=\textwidth]{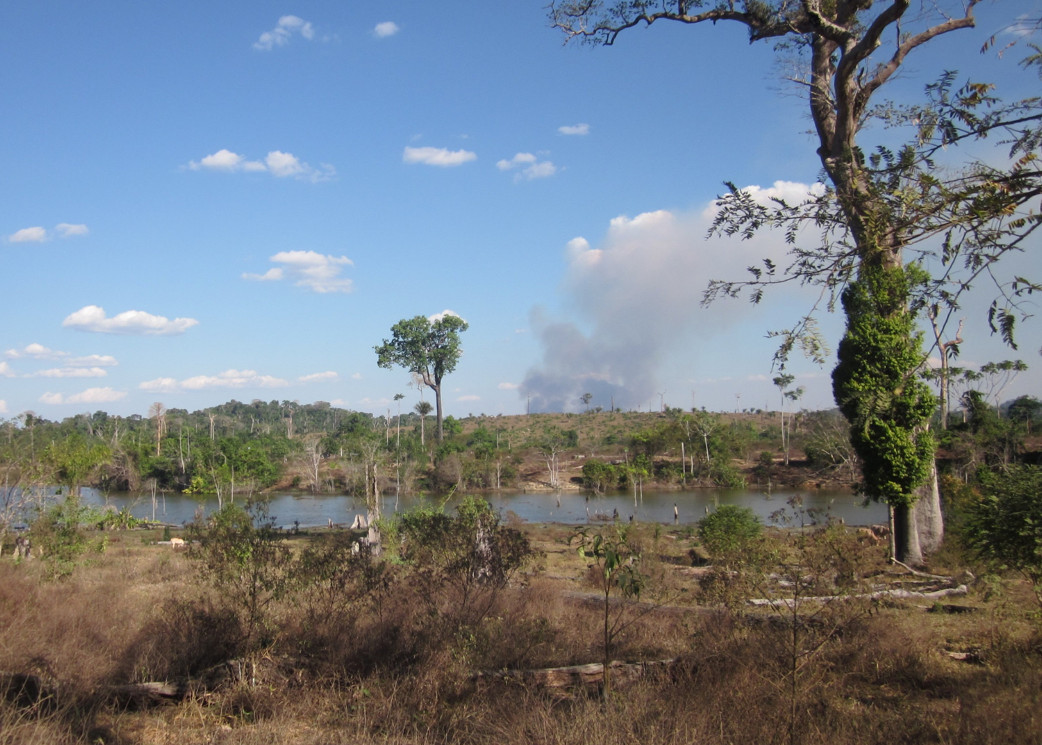}
\caption{\label{fig:foto}Photograph illustrating the fluent transitions and interactions of different land cover types in the study region.}
\end{figure}
\begin{table}
	\centering
	\caption{\label{tab:scenes}Scenes utilized in this study. All scenes were collected over the same area using the same incidence angle.}
	\begin{tabular}{llr} \hline
		Date & Polarization\\ \hline
		2014-06-08 & VV-VH \\
		2014-06-30 & HH-HV \\
		2014-07-22 & VV-VH \\
		2014-08-24 & VV-VH \\
		2014-09-04 & HH-HV \\ \hline
	\end{tabular}
\end{table}
\subsection{Remote Sensing Data}\label{prep}
The data base for the study consists of five TS-X strip map scenes with $5m \times 5m$ spatial resolution (Table \ref{tab:scenes}). All images are ordered in single-look complex format, comprising different VV-VH and HH-HV polarization at an incidence angle of $37.75^{\circ}$, and cover a swath of roughly $50 km \times 15 km$ ($5663 \times 11856$ pixels).
Data is calibrated and processed according to common procedures (see Section \ref{prep}). Preprocessing in the context of this study includes all necessary steps before random sampling of training and test data is performed. After random sampling, training of IVM, and MRF regularization are taken out, land cover maps are generated and validated, and average measures are calculated.\\Preprocessing of the TS-X scenes is conducted using the Sentinel 1-Toolbox and the Geospatial Data Abstraction Library \citep{GDAL}. All scenes were processed separately in the following order:
\begin{description}
	\item[Multilooking:]3 range looks, 2 azimuth looks, yielding a ground resolution of $\sim 4.7m \times 4.7m$.
	\item[Terrain \& Radiometric Correction:]Range-Doppler terrain correction (SRTM 3Sec) and resampling to $5m \times 5m$ pixel spacing. The data is projected into UTM Zone 21S. $\gamma^0$ radiometric normalization is applied using an SRTM.
	\item[GLCM-Texture]Texture measurements are widely used to increase the mapping accuracy of SAR data \citep{sarker2012potential,dekker2003texture,cutler2012estimating}. Gray Level Co-occurence Matrices (GLCM) are calculated with $11 \times 11$ moving window size into all symmetric directions with offset one. Probabilistic quantization is conducted into 64 levels.  Ten texture parameters are separately derived for any available polarization and any available scene: contrast, dissimilarity, homogeneity, angular second moment (ASM), energy, maximum probability, entropy, GLCM mean, GLCM variance, and GLCM correlation. These will be included as additional features to improve the IVM classification. For more information on GLCM-based texture parameters see \citet{haralick1973textural,sarker2012potential}. In respect to findings by \citet{sarker2013forest,Nyoungui2002}, and our own experiments, we abstain from combining texture metrics with speckle filtering. Since we use 10 texture measures per layer, we have a total of 22 features per scene for the classification process.
\end{description}
\subsection{Reference Data}\label{subsec:ref}
Reference data includes multispectral RapidEye and Landsat data, in situ data, as well as land cover data from various Brazilian agencies (e.g. PRODES, TerraClass). PRODES (Programa de C\'{a}lculo do Desflorestamento da Amaz\^{o}nia) is an effort by the Brazilian space agency (INPE) to generate annual maps documenting deforestation of primary forests inside the Legal Amazon with a minimum mapping unit of $6.25ha$ \citep{inpe2015projetoprodes}. Targeting only the sites that PRODES considers deforested, TerraClass is an effort to determine LULC classes of the affected areas \citep{ALMEIDA2016}. The overall coverage of all available TS-X swaths constitutes the study area (Figure \ref{fig:tsxsaarea}), and is sufficiently covered by reference information. While forests as well as clean and shrubby pasture are present in the study area, occurrence of water and burnt pasture is overall scarce. To address this issue, polygons are manually distributed over the entire area. Afterwards, each polygon is assigned one class label for each date covered by TS-X to address changes of LULC. If necessary, polygons are split to avoid class ambiguity within different temporal instances. E.g., if a coherent pasture area is only partially burnt, the polygon gets split. The generation of reference data is supported by visual interpretation of RapidEye as well as Landsat 7 ETM+ and Landsat 8 OLI data of the same time period. In addition to Landsat and RapidEye imagery, fire products derived from MODIS are also considered. Moreover, photographs from a field campaign conducted in August 2014 are available.
\begin{table}
\centering
	\small
	\caption{\label{tab:classes}Number of sample points available for training distinguished by class, extracted from polygons.}
	\begin{tabular}{lrrrrr}
		\toprule
        & \multicolumn{5}{c}{Date}\\
        \midrule
		Class & 06-08 & 06-30 & 07-22 & 08-24 & 09-04 \\
		\midrule
		Burnt Pasture & 15 & 55 & 139 & 469 & 434 \\
        Clean Pasture & 487 & 444 & 375 & 384 & 420 \\
        Shrubby Pasture & 749 & 783 & 789 & 438 & 428 \\
        Water & 28 & 28 & 28 & 28 & 28 \\
        Forest & 799 & 701 & 704 & 616 & 516 \\
		\bottomrule
	\end{tabular}
\end{table}
\\Sampling is conducted by two of the authors in close cooperation and was harmonized with classification schemes by INPE (Instituto Nacional de Pesquisas Espaciais). The following LULC classes are considered:
\texttt{clean pasture, shrubby pasture, burnt pasture, water,} and \texttt{forest}.
\texttt{Clean pasture}, also called pasto limpo, describes pasture land that is intensively worked.
This includes regular tillage and burning of land to support cattle ranching.
\texttt{Shrubby pasture}, also called pasto sujo, is not intensively managed and thus affected by bush encroachment.
The coarser appearence of \texttt{shrubby pasture} generally allows a visual separation from \texttt{clean pasture} in high resolution images.
\texttt{Burnt pasture} includes clean as well as shrubby pasture areas which were recently burned, and are characterized by open soil and vegetation residues. Such areas can be easily identified using false color composites.
\texttt{Forest}, beside primary forests, includes areas of secondary vegetation and regeneration as these are usually non-separable by X-band SAR.
\begin{table}
	\small
	\begin{tabular}{lcc}
		\toprule
        Class & In-Situ & Multispectral (RE, LS) \\
        \midrule
         \rotatebox[origin=l]{90}{\makebox[1in]{Pasto Sujo}} & \includegraphics[width=0.4\textwidth,height=0.26\textwidth]{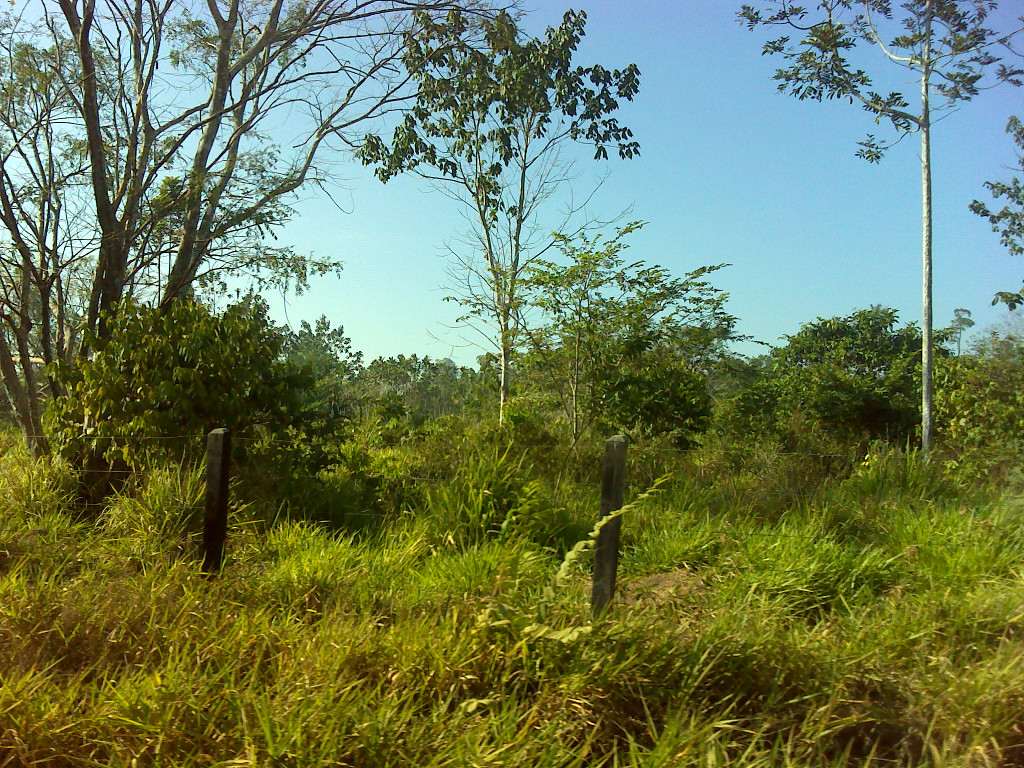} & \includegraphics[width=0.4\textwidth,height=0.26\textwidth]{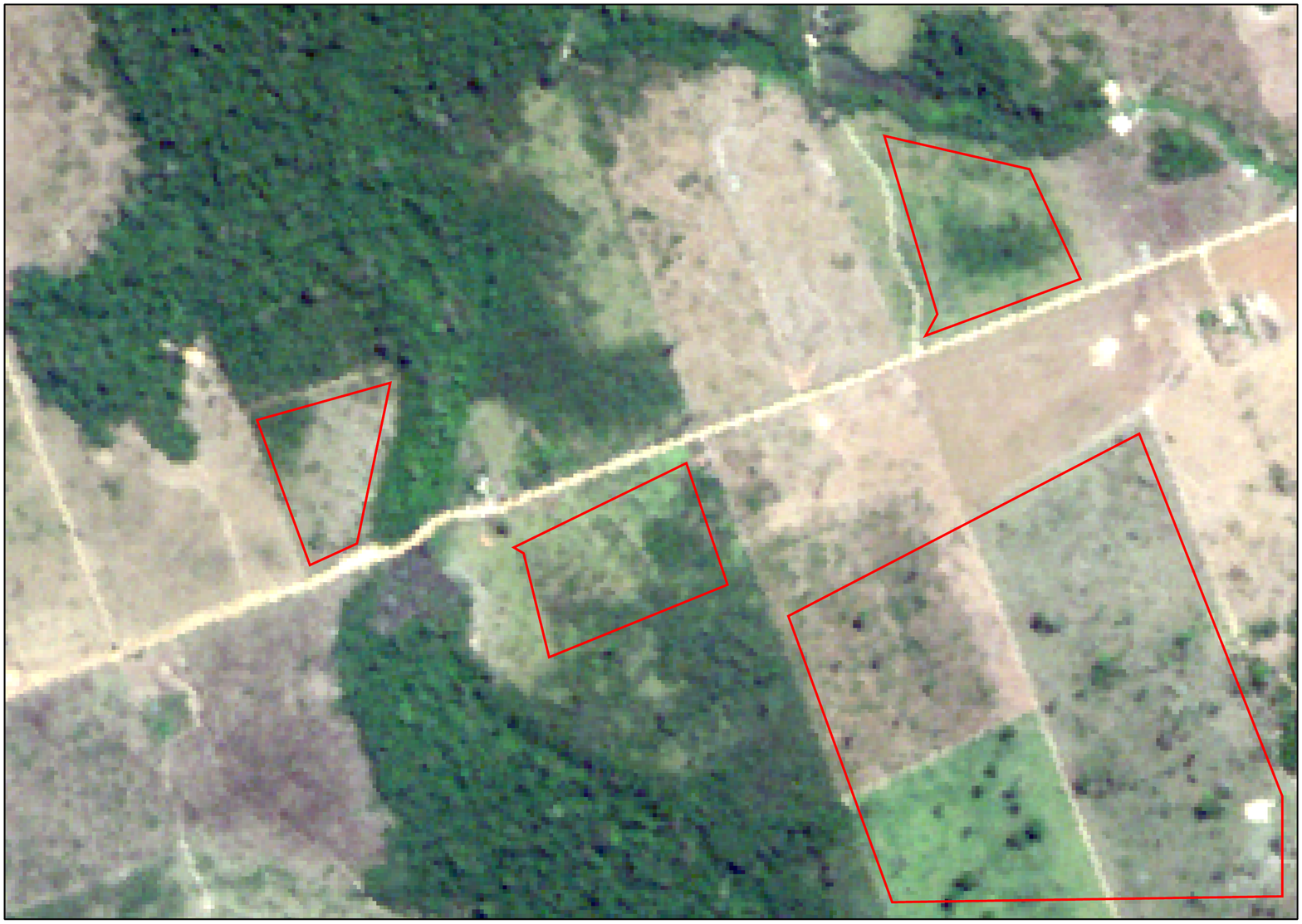} \\
        \rotatebox[origin=l]{90}{\makebox[1in]{Pasto Limpo}} & \includegraphics[width=0.4\textwidth,height=0.26\textwidth]{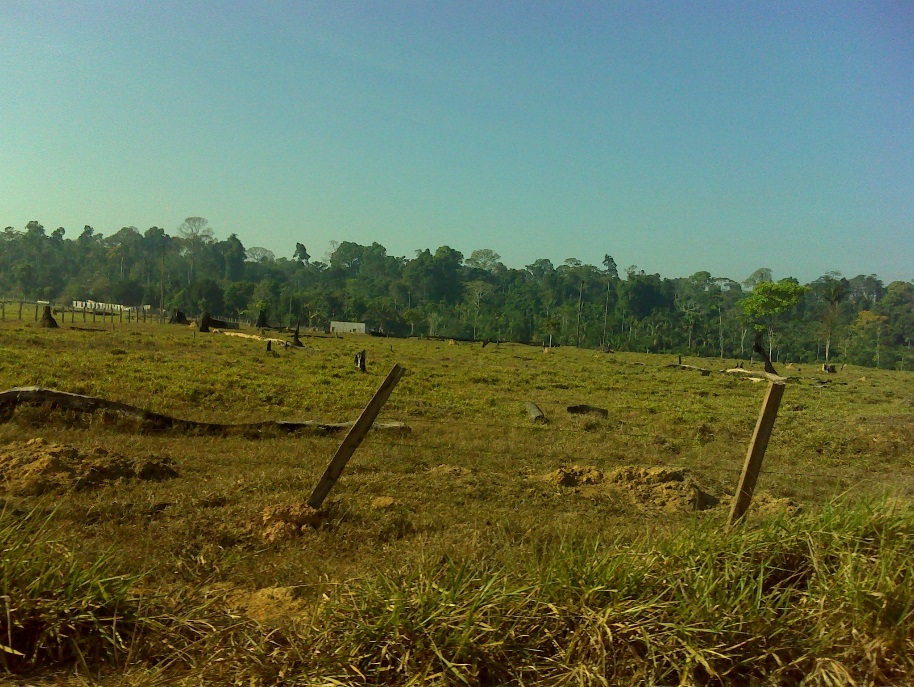} & \includegraphics[width=0.4\textwidth,height=0.26\textwidth]{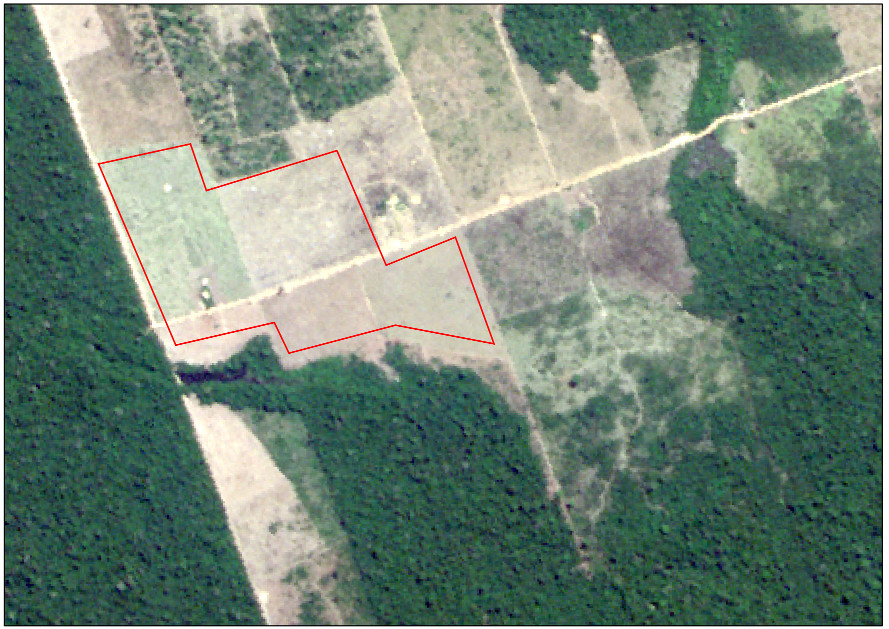} \\
        \rotatebox[origin=l]{90}{\makebox[1in]{Pasto Burnt}} & \includegraphics[width=0.4\textwidth,height=0.26\textwidth]{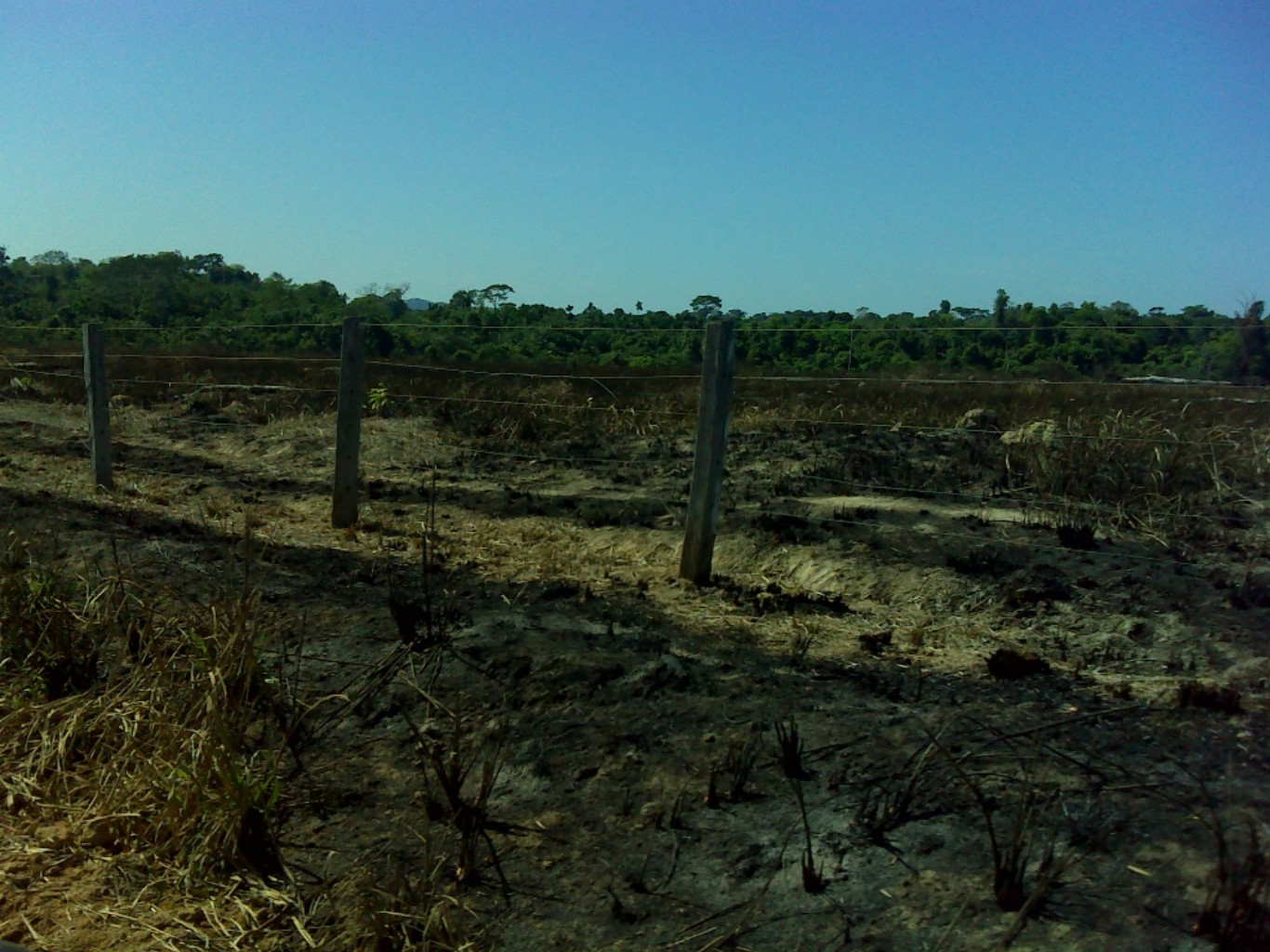} & \includegraphics[width=0.4\textwidth,height=0.26\textwidth]{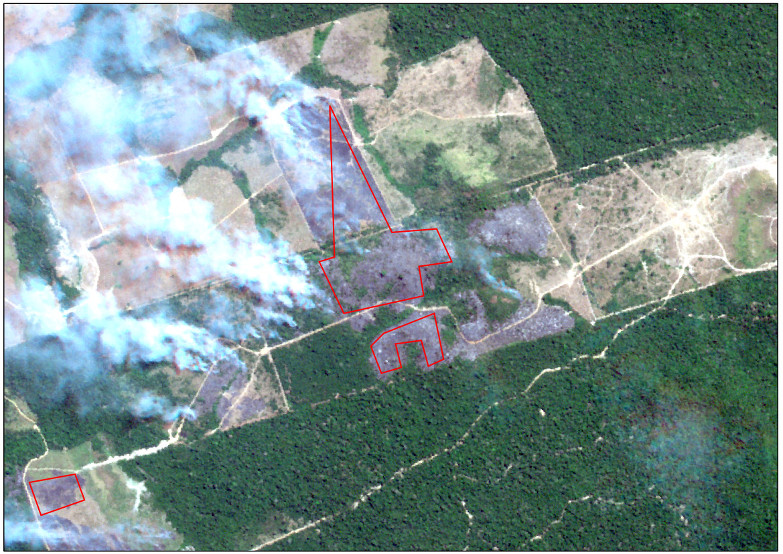} \\
        \rotatebox[origin=l]{90}{\makebox[1in]{Forest}} & \includegraphics[width=0.4\textwidth,height=0.26\textwidth]{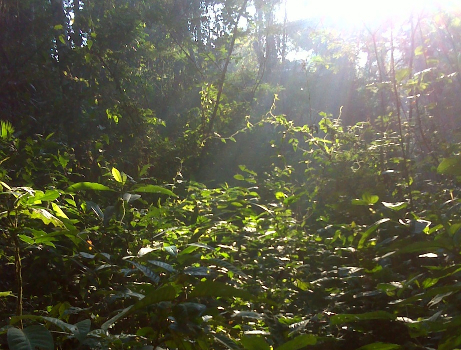} & \includegraphics[width=0.4\textwidth,height=0.26\textwidth]{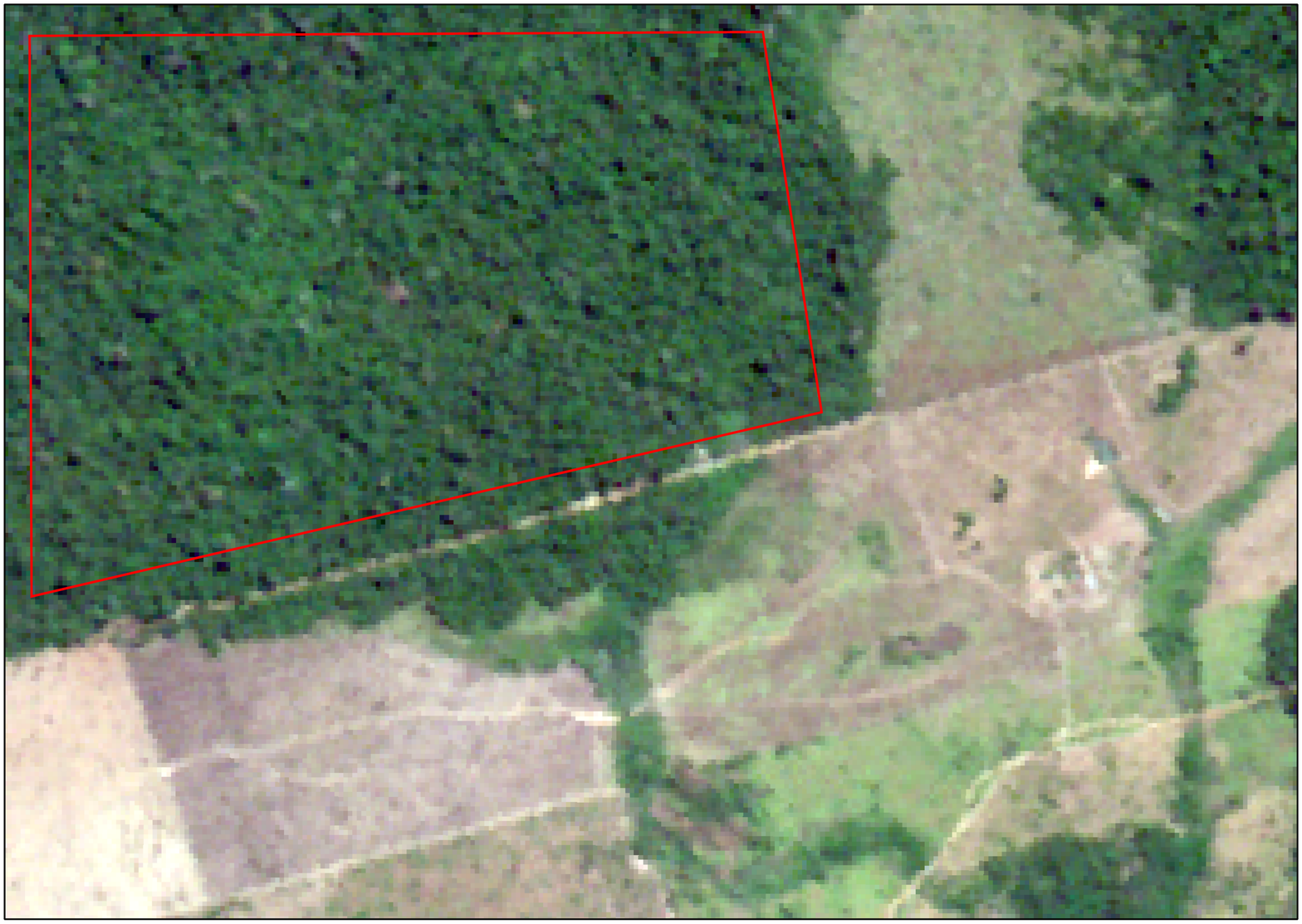} \\
        \rotatebox[origin=l]{90}{\makebox[1in]{Water}} & \includegraphics[width=0.4\textwidth,height=0.26\textwidth]{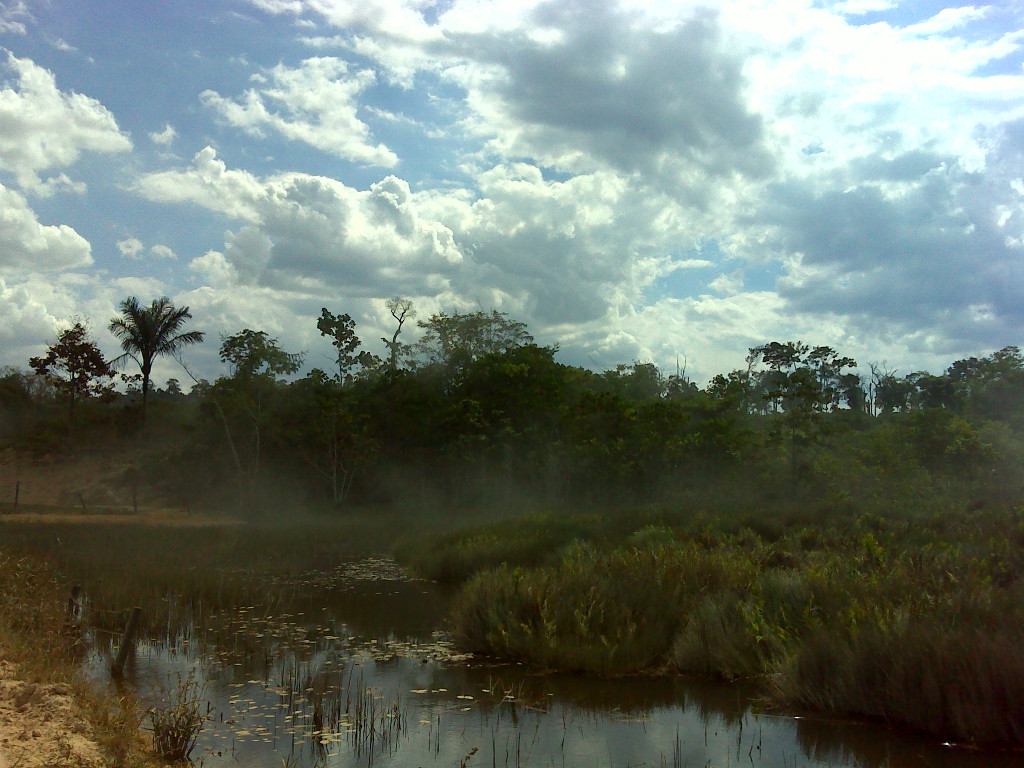} & \includegraphics[width=0.4\textwidth,height=0.26\textwidth]{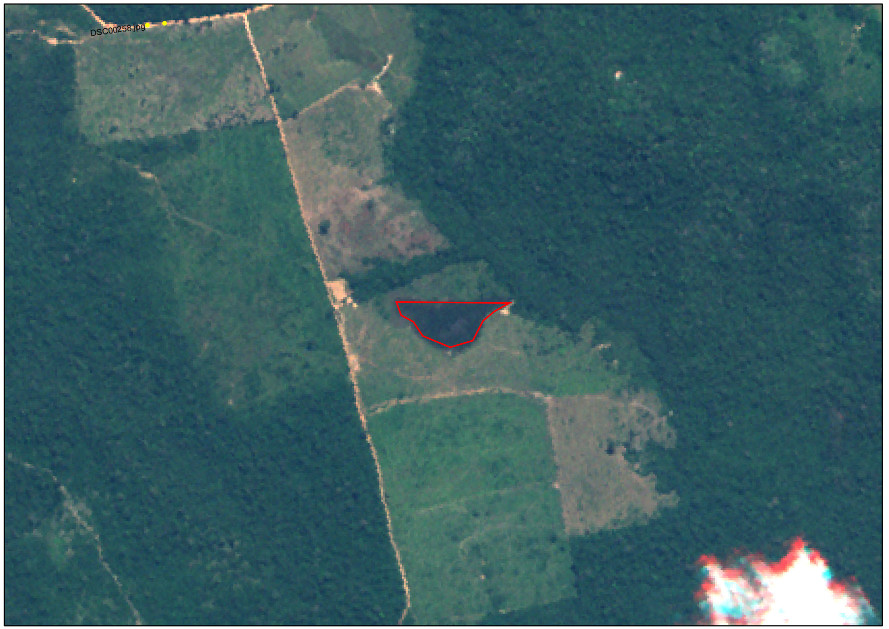} \\
		\bottomrule
	\end{tabular}
    \captionof{figure}{Classification scheme.} \label{fig:classkey}
\end{table}
Forests have a very characteristic appearance in TS-X images and high resolution multispectral imagery.
Table \ref{tab:classes} gives an overview of the number of available training samples for each class and date.
It should be underlined that the burning season usually starts around end of July.
Hence only few burned pasture areas could be identified before that period.
Water bodies are also very scarce and only two lakes over the entire study area are included.
Figure \ref{fig:classkey} visualizes the classes considered in our classification scheme.
The considered LULC classes match comparable studies using TS-X data in Brazilian, or tropical settings, respectively \citep{garcia2011analise,schlund2013importance}. As the time period of our study falls into the dry season between June and September, corresponding multispectral remote sensing data could be interpreted sufficiently well. Yet, some challenges remain:
\begin{itemize}
\item For the study region two dominating pasture types can be identified: pasto sujo, i.e., \texttt{shrubby pasture}, and pasto limpo, i.e., \texttt{clean pasture} \citep{almeida2008dinamica,adami2015dinamica}. While both types are generally used for cattle ranching in this region, pasto sujo is characterized by bushes and occasional early stages of succession.
However, the transition between these types can be gradual and consequently hard to interpret from remote sensing imagery alone; even at $5 m$ ground resolution as offered by RapidEye.
\item Transitions from pasto sujo into early stages of secondary vegetation are hard to distinguish, due to the gradual nature of this process. However, it is not as relevant in our study site since significantly less areas are affected. To allow a solid separation of these classes, we consider multi-annual time series to identify pasture management. In addition, we include information offered by TerraClass which reliably separate different types of secondary vegetation and pasture land.
\item Primary and secondary forests, as well as secondary vegetation, are combined into one class, as various studies and own preliminary tests indicate the limitation of X-band for separating these two classes.
\end{itemize}
\section{Methods}\label{methods}
The proposed framework consists of four steps: (i) preprocessing, (ii) random sampling, (iii) classification of each single scene using IVM, and (iv) optimization of the MRF model. Final validation is performed on averages over 10 independent runs, using a random sampling (50:50) into spatially disjoint train and test polygons. As pixels sampled from training polygons are solely used for IVM parameterization (grid search) and model training, pixels sampled from test polygons enable an independent validation.
\\Throughout the paper we use the following notation: Let there be a training set $\(\d x_n, y_n\) \in \mathcal T$ comprising $N$ feature vectors $\{\d x_1, \ldots, \d x_n, \ldots, \d x_N\}$ and corresponding class labels $y_n \in \{1,...,K\}$, distributed over an image lattice $\m I$. 
We later address image samples at any given coordinate as $\boldsymbol{x}_i$, and probability estimates as $\boldsymbol{p}_i=[p_1,...,p_{nk},...,p_{nK}]$ with $p_{nk}= p(y_{n}=k\mid \d x_{n})$.
\subsection{Import Vector Machines}
\label{ivm}
IVM is a discriminative and probabilistic classifier based on kernel logistic regression and has first been introduced by \cite{zhu2005kernel}. 
\cite{roscher2012incremental} have shown that IVMs provide more reliable probabilities than probabilistic SVMs, since IVMs' probabilities are more balanced, whereas SVMs generally overestimate maximum probabilities.
To account for complex decision boundaries between classes, IVM generally benefit from integrating a kernel function. 
For this study, we utilize the radial basis function (rbf) kernel parameterized by kernel width $\sigma$, which is a standard for remote sensing purposes.
Parameterization is achieved analogously to standard SVM practices using a grid search, to estimate the cost parameter $C$ and $\sigma$.
For a more encompassing description of IVM see \citet{zhu2005kernel} and \citet{roscher2012sup}.
\subsection{Markov Random Field}
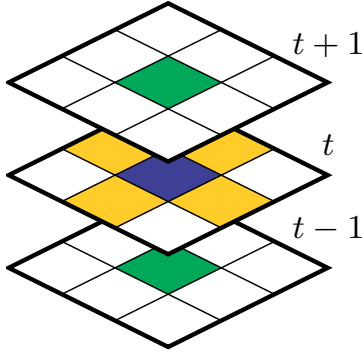
\begin{figure}
				\centering
				\small
                \begin{tikzpicture}[scale=.5,every node/.style={minimum size=1cm},on grid]	
				\draw[anchor=west] (2,9.9) node {$t+1$};
				\draw[anchor=west] (2,8) node {$t$};
				\draw[anchor=west] (2,6.4) node {$t-1$};	
				\begin{scope}[
				yshift=90,every node/.append style={
					yslant=0.5,xslant=-1},yslant=0.5,xslant=-1
				]
				\fill[white,fill opacity=.4] (1,1) rectangle (4,4);
				\fill[green] (2,2) rectangle (3,3);

				\draw[step=10mm, black] (1,1) grid (4,4);
			\draw[black,very thick] (1,1) rectangle (4,4);
		
				\end{scope}
				\begin{scope}[
				yshift=140,every node/.append style={
					yslant=0.5,xslant=-1},yslant=0.5,xslant=-1
				]
				\fill[white] (1,1) rectangle (4,4);
											\fill[blue] (2,2) rectangle (3,3);
											\fill[yellow] (2,3) rectangle (3,4);
											\fill[yellow] (3,2) rectangle (4,3);
											\fill[yellow] (1,2) rectangle (2,3);
											\fill[yellow] (2,1) rectangle (3,2);
				\draw[step=10mm, black] (1,1) grid (4,4);
				\draw[black,very thick] (1,1) rectangle (4,4);
		
				\end{scope}
				\begin{scope}[
				yshift=190,every node/.append style={
					yslant=0.5,xslant=-1},yslant=0.5,xslant=-1
				]
				\fill[white] (1,1) rectangle (4,4);
				\fill[green] (2,2) rectangle (3,3);
				\draw[step=10mm, black] (1,1) grid (4,4);				
				\draw[black,very thick] (1,1) rectangle (4,4);
			
				\end{scope}
				
				\end{tikzpicture}
				\caption{\label{fig:neighbors}Temporal (green) and spatial (yellow) neighbors of a given pixel (blue).}
\end{figure}
In this study, we use post classification MRF with spatio-temporal neighborhood relations between pixels, as illustrated in Figure \ref{fig:neighbors}. Parameterization is achieved through transition matrices, which are $5 \times 5$ matrices indicating spatial and temporal transition probabilities between the five classes. For our description of MRF, we adapt a terminology similar to \citet{moser2013land,melgani2003markov}. 
Therefore, with $\boldsymbol x_i$ denoting pixel features and $y_i$ its corresponding class label, we reformulate the IVM-based probabilities $p(y_{i}\mid \boldsymbol x_i)$ as energy terms
\begin{equation}\label{eq:specenergy}
U_{\m X} = - \sum_{i\in I}\ln p(y_{i}\mid \boldsymbol x_{i}).
\end{equation}
As the energy $U_{\m X}$ are equivalent, minimization of $U_{\m X}$ is identical to maximization of $P$.
Now consider a function for the spatial neighborhood $U_{\text{sp}}$, with $i\sim_\text{sp} j$ applying to any two pixels, which are direct, 4-connected, spatial neighbors, and a function $\delta(y_i,y_j)$ to assign weights to neighboring classes:
\begin{equation}\label{eq:spatialenergy}
U_{sp} = \sum_{i\sim_s j}1-\delta(y_{i},y_{j})
\end{equation}
In this case the function $\delta$ yields a $K \times K$ matrix which can be used to favor certain neighboring constellations. The function $\delta$ is generally defined as Potts model to result in an identity matrix, which encourages the generation of homogeneous areas. 
The standard mono-temporal MRF model is given by summation of \eqref{eq:specenergy} and \eqref{eq:spatialenergy}
\begin{equation}\label{eq:specspate}
U_{\m X,\text{sp}} = - \sum_{i\in I}\ln p(y_{i}\mid \d x_{i}) + \beta \sum_{i\sim_s j}1-\delta(y_{i},y_{j})\;,
\end{equation}
 with weight parameter $\beta$ to regulate importance of the spatial component.
\label{MTMRF}
For the multi-temporal case we consider co-registered images with the temporal neighbors being the spatially congruent cells at the neighboring acquisition times $t-1$ and $t+1$. 
Only if pixel $j$ is the temporal successor of $i$, $i\sim_{t+1}j$ applies; and only if pixel $j$ is the temporal predecessor of $i$, $i\sim_{t-1}j$ applies. 
Temporal energy is hence given by \eqref{eq:temporalenergy}, analogous to the spatial case.
\begin{equation}\label{eq:temporalenergy}
U_{\text{temp}} = \sum_{i\sim_{t+1}j}1-\tau_1(y_{i},y_{j})+\sum_{i\sim_{t-1}j}1-\tau_2(y_{i},y_{j})
\end{equation}
Here, $\tau_1$ and $\tau_2$ are $K \times K$ matrices defining the temporal transitions as observed from land cover trajectories. In opposition to the spatial weighting $\delta$, we require multiple, non symmetrical matrices $\tau$ to respect trajectories with regard to the future, or the past. The overall energy function is defined by integrating the temporal vicinity into \eqref{eq:specspate}, which yields
\begin{equation}\label{eq:fullenergy}
\begin{split}
U = - \sum_{i\in I}\ln p(y_{i}\mid \d x_{i}) + \beta_{sp} \sum_{i\sim_s j}1-\delta(y_{i},y_{j}) \\
+ \beta_{\text{temp}}\Big(\sum_{i\sim_{t+1}j}1-\tau_1(y_{i},y_{j})+\sum_{i\sim_{t-1}j}1-\tau_2(y_{i},y_{j})\Big)\;.
\end{split}
\end{equation}
This function combines \eqref{eq:specspate} and \eqref{eq:temporalenergy}. 
Weight parameters $\beta_{(\cdot)}$ can be used to adjust the importance of temporal and spatial weights.
\subsection{Passing scheme \& transition matrices}
\label{opt}
\begin{figure}
	\centering
	\begin{tikzpicture}[node distance=1cm,auto]	
	\tikzstyle{varcenter}=[rectangle,thick,draw=black,fill=blue,minimum size=7mm]
	\tikzstyle{varspat}=[rectangle,thick,draw=black,fill=yellow,minimum size=7mm]
	\tikzstyle{vartemp}=[rectangle,thick,draw=black,fill=green,minimum size=7mm]
	\tikzstyle{factorspat}=[circle,thick,draw=black,minimum size=7mm]
	\tikzstyle{factortemp}=[circle,thick,draw=black,minimum size=7mm]
	\tikzstyle{factorun}=[circle,thick,draw=black,minimum size=7mm]
	\tikzstyle{every label}=[red]
	\begin{scope}
	\node [varcenter] (P) {};
	\node [factorspat] (db) [below=1cm of P] {$\small\delta$};
	\node [factorspat] (da) [above= of P] {$\small\delta$};
	\node [factorspat] (dl) [left= of P] {$\small\delta$};
    \node [factorspat] (dr) [right=of P] {$\small\delta$};
    \node [factortemp, position=-30:1cm from P] (tbr) {$\small\tau_1$};
    \node [factortemp, position=-60:1cm from P] (tbr2) {$\small\tau_2$};
    \node [factortemp, position=150:1cm from P] (tal) {$\small\tau_1$};
    \node [factortemp, position=120:1cm from P] (tal2) {$\small\tau_2$};
	\node [factorun] (ivm) [above right= of P] {\small I};
	\node [varspat] (L) [left=of dl] {};
	\node [varspat] (R) [right=of dr] {};
	\node [varspat] (B) [below=of db] {};
	\node [varspat] (A) [above=of da] {};
	\node [vartemp, position=135:2cm from P] (T1) {};
    \node [vartemp, position=-45:2cm from P] (T-1) {};
	\draw [-] (P.west) -- (dl.east);
	\draw [-] (P.north) -- (da.south);
	\draw [-] (P.south) -- (db.north);
	\draw [-] (P.east) -- (dr.west);
	\draw [-] (P.north east) -- (ivm.south west);
	\draw [-] (P.west) -- (tal.south east);
    \draw [-] (P.north) -- (tal2.south east);
	\draw [-] (P.east) -- (tbr.north west);
    \draw [-] (P.south) -- (tbr2.north west);
	\draw [-] (dl.west) -- (L.east);
	\draw [-] (dr.east) -- (R.west);
	\draw [-] (da.north) -- (A.south);
	\draw [-] (db.south) -- (B.north);
	\draw [-] (tal.north west) -- (T1.south);
    \draw [-] (tal2.north west) -- (T1.east);
	\draw [-] (tbr.south east) -- (T-1.north);
    \draw [-] (tbr2.south east) -- (T-1.west);
	\end{scope}
	\end{tikzpicture}
	\caption{\label{fig:snowflake}Factor-graph as implemented in this study. Variable nodes illustrated by yellow (spatial neighbor) and green nodes (temporal). Circles mark the corresponding factor nodes and the unary IVM-based energy.}
\end{figure}
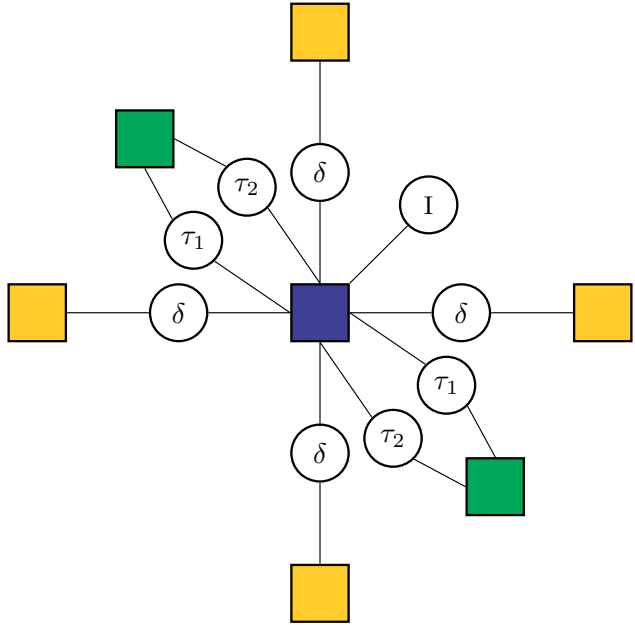

LBP is an inference algorithm utilizing Message Passing \citep{pearl1982reverend}, and is shown to approximate maximum values sufficiently well \citep{murphy1999loopy}. We choose LBP over graph-cut based methods for their more general applicability, as graph-cuts are specifically defined for symmetrical binary factors \citep{boykov2001fast}, and can not be applied in non-symmetric environments \citep{kolmogorov2004energy}.
ICM (Iterated Conditional Modes) is another algorithm which is commonly used to achieve inference, especially in remote sensing and using multi-temporal data sets \citep{liu2008using,liu2006spatial,wehmann2015spatial}. While it has low computational cost, it is generally outperformed by LBP in terms of accuracy \citep{Szeliski2006,andres2010empirical}. For this reason we formulate an implementation of LBP using moving windows, which can be applied to image stacks of arbitrarily large image stacks sufficiently well. 
Figure \ref{fig:snowflake} illustrates the neighborhood of one pixel in a factor graph, analogous to the MRF neighborhood as described in Section \ref{MTMRF}.
\begin{table}
\begin{tabular}{m{0.4\textwidth}m{0.45\textwidth}}
\multicolumn{1}{c}{\textbf{Action}} & \multicolumn{1}{c}{\textbf{Description}} \\ \toprule
\includegraphics[width=0.4\textwidth]{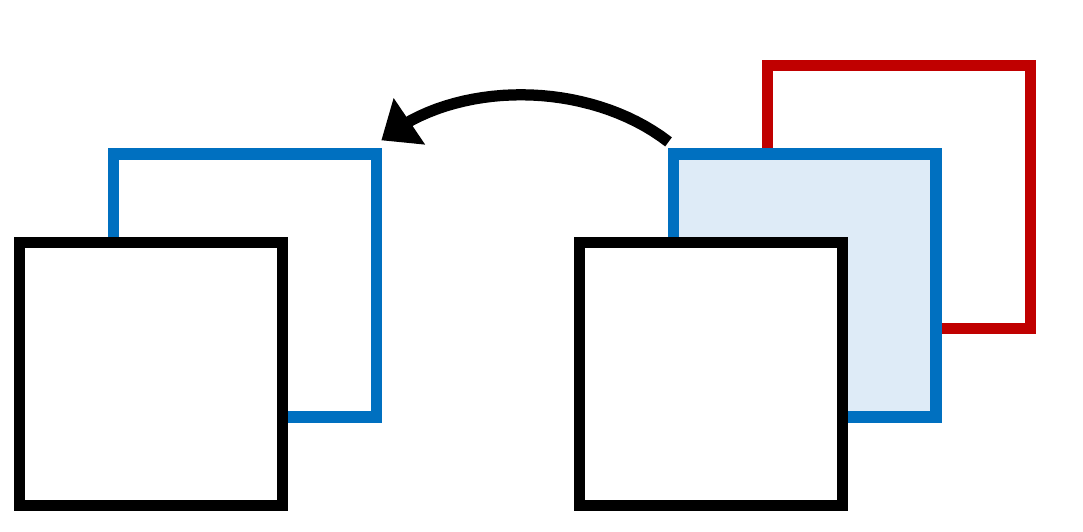} &\textbf{Step 1:} Generation of a fallback copy of the current energy layer (blue). Necessary for future calculation of messages passed to the red layer.\\
\includegraphics[width=0.4\textwidth]{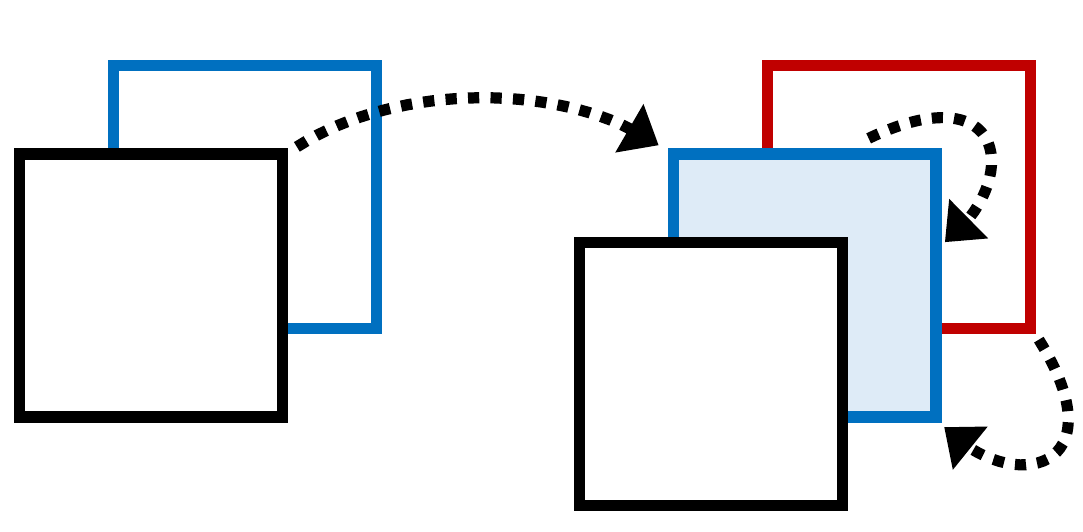} &\textbf{Step 2:} Messages are passed from the previous fallback, the next energy layer, and the current layer, also factoring in the unaries upon receiving. This procedure is performed using a set of moving windows.\\
\includegraphics[width=0.4\textwidth]{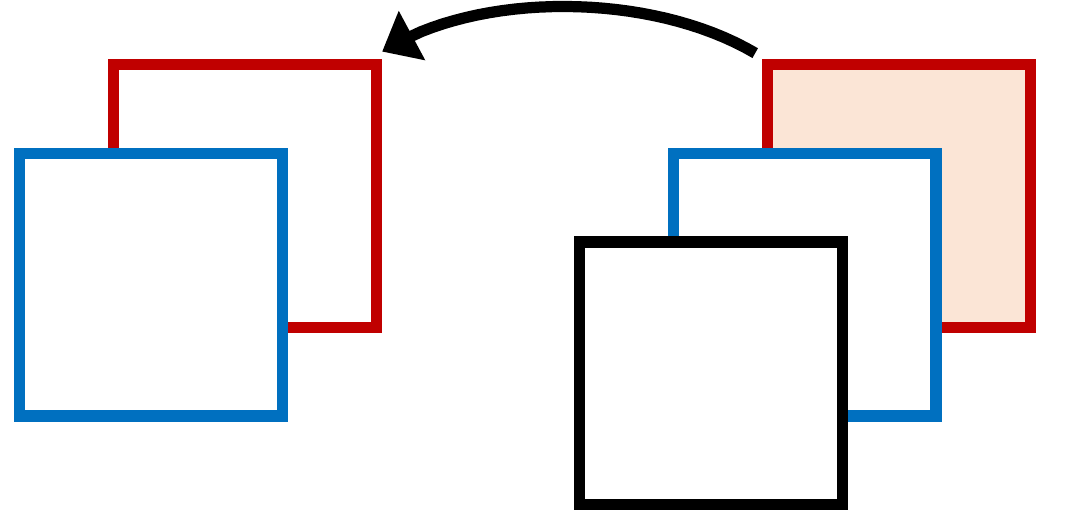} &\textbf{Step 3:} Discarding of the previous fallback, and backing up of the next energy layer (see Step 1). Iterate over each layer.\\ \bottomrule
\end{tabular}
\captionof{figure}{Passing schedule as applied in this study. One pass over all layers corresponds to one iteration of LBP.} \label{fig:steps}
\end{table}
Using the Potts function to define $\delta$ is common practice in remote sensing literature \citep{moser2013land}, and since the focus of this study lies on the examination of MRF for a multi-temporal linking of classifications, we follow this practice. The Potts function can be represented by an identity matrix, which supports assignment of neighboring pixels to the same class. It is in general not sensible to formulate an asymmetric message passing for the two spatial dimensions, as a pixel will assume the same properties of its left as of its right neighbor. More specifically, the Potts model is a way to reflect Tobler's assumption on spatial autocorrelation, promoting the idea of close objects to be more alike than distant objects.\\In contrast to the spatial transitions, utilization of the Potts model for temporal transitions can lead to serious distortions and cause equalization of all subjected probability maps, i.e., it would prohibit any land cover changes from occuring. While we can not assume any spatially directional patterns in the area, and thus rely on the Potts model, a pixel will pass different messages to its temporal successor opposed to its predecessor. Through adjustment of $\tau$ it is possible to assign probabilities for all possible types of class transitions. We therefore express temporal transitions through two asymmetric transition matrices ($\tau_1$ and $\tau_2$). While matrix $\tau_1$ illustrates the messages pixels pass from the scene at time $t$ to its neighbor at $t-1$, $\tau_2$ defines the messages the pixels pass from $t$ to $t+1$. This differentiation is important, e.g. considering that burnt pasture at $t$ will prohibit primary forests at $t+1$, yet it might endorse pasture at $t-1$.
We integrate the user-defined transition matrices $\tau_1$ and $\tau_2$ as interface to inject a-priori expert knowledge into our regularization model. While not empirically derived, these matrices are based on weak assumptions on land cover trajectories.
\\The assumptions include that \texttt{Water} and \texttt{Forest} are regarded very consistent classes, yet that pasture areas have some kind of interaction with each other. This especially concerns the transition of pasture land to \texttt{burnt pasture} land, which is explicitly tolerated. Furthermore, \texttt{forest} does not explicitly prohibit predecessing non-forest areas, which is due to \texttt{forest} including secondary vegetation and to offer the model some tolerance with regard to misclassifications. Hence, the formalization of land cover trajectories is relatively straight forward, and not necessarily based on elaborate a-priori knowledge. Previous tests showed a very similar outcome concerning the modification of these parameters, yet using very strict transitions could lead to undesired results and suppress dynamics entirely.
In the scope of this study we utilize different transition matrices due to the varying time gap between the five TS-X acqusitions. The revisit rate of TS-X is eleven days, and the available imagery shows one gap of eleven days, two gaps of 22 days, as well as one gap of 33 days in between neighboring acquisitions. We hence linearly modify the transitions to adjust to the varying temporal resolution, since with increasing time, more change is expected to occur.
\\The following summarizes the most relevant assumptions we made for specification of the transition matrices.
\begin{itemize}
	\item Pasture areas can potentially be burnt. After burning, likelihood is high to transition back into \texttt{clean pasture} or \texttt{shrubby pasture}.
	\item Transitions from \texttt{shrubby pasture} to \texttt{clean pasture} are permitted.
	\item \texttt{Clean pasture} is considered stable, yet may transition into \texttt{shrubby pasture} or \texttt{forest}. Following observations of TerraClass, for the study region we assume a slow shift from \texttt{clean pasture} into \texttt{shrubby pasture} overall. 
	\item \texttt{Forest} is the most consistent class. It can get removed, yet especially \texttt{shrubby pasture} can develop into \texttt{forest} as the class also includes secondary vegetation.
	\item \texttt{Water} is used to describe bodies of water which are permanently filled within this dry season.
	\item For any class there is a small tolerance to evade to counteract inconsistent transitions which may be caused by misclassifications.
\end{itemize}
\subsection{Classification \& Validation}
\label{subsec:train}
Three different types of classification are compared: (i) the baseline IVM classification, (ii) the spatial-only MRF with $\beta_t=0$, from now on referred to as s-MRF, and (iii) the spatial-temporal MRF, referred to as st-MRF. While many studies rely on supervised classification using SVM and Random Forest classifiers, various studies show that IVM perform at least equally well in terms of accuracy (\citep{roscher2012incremental,Braun2012}). Therefore the original IVM classification is considered as adequate baseline classification.\\
Reference polygons exclusively comprise either training or test samples to avoid spatial autocorrelation.
For training purposes, $15$ samples per polygon are randomly selected, using a minimum sampling distance of $30$~meters.
A systematic sampling ensures that an adequate number of training samples is selected for all five classes; \texttt{clean pasture, shrubby pasture, burnt pasture, water,} and \texttt{forest}.
Validation is conducted considering the current terms of good practice as laid out by \citet{Olofsson2014}. 
Samples are clustered in polygons to improve on the spatial variability of both, training and test samples, with pixels being the assessment unit.
This sampling strategy is a necessary trade-off between ideal conditions of independent random sampling and the difficulties of obtaining large scale, multi-temporal reference data in a challenging environment \citep{Olofsson2014}.
Error matrices are derived to serve as a basis for the estimation of overall accuracies (OA), user accuracies (UA), producer accuracies (PA), and their corresponding confidence intervals (CI). 
In addition, we calculate area measures and their confidence intervals at each acquisition date to estimate the development of \texttt{burnt pasture} land over the entire 2014's dry season. 
Classification and validation is conducted ten times using different training and test sets and the results are averaged.
\section{Results}\label{result}
\begin{figure}
	\centering
	\includegraphics[width=0.8\textwidth]{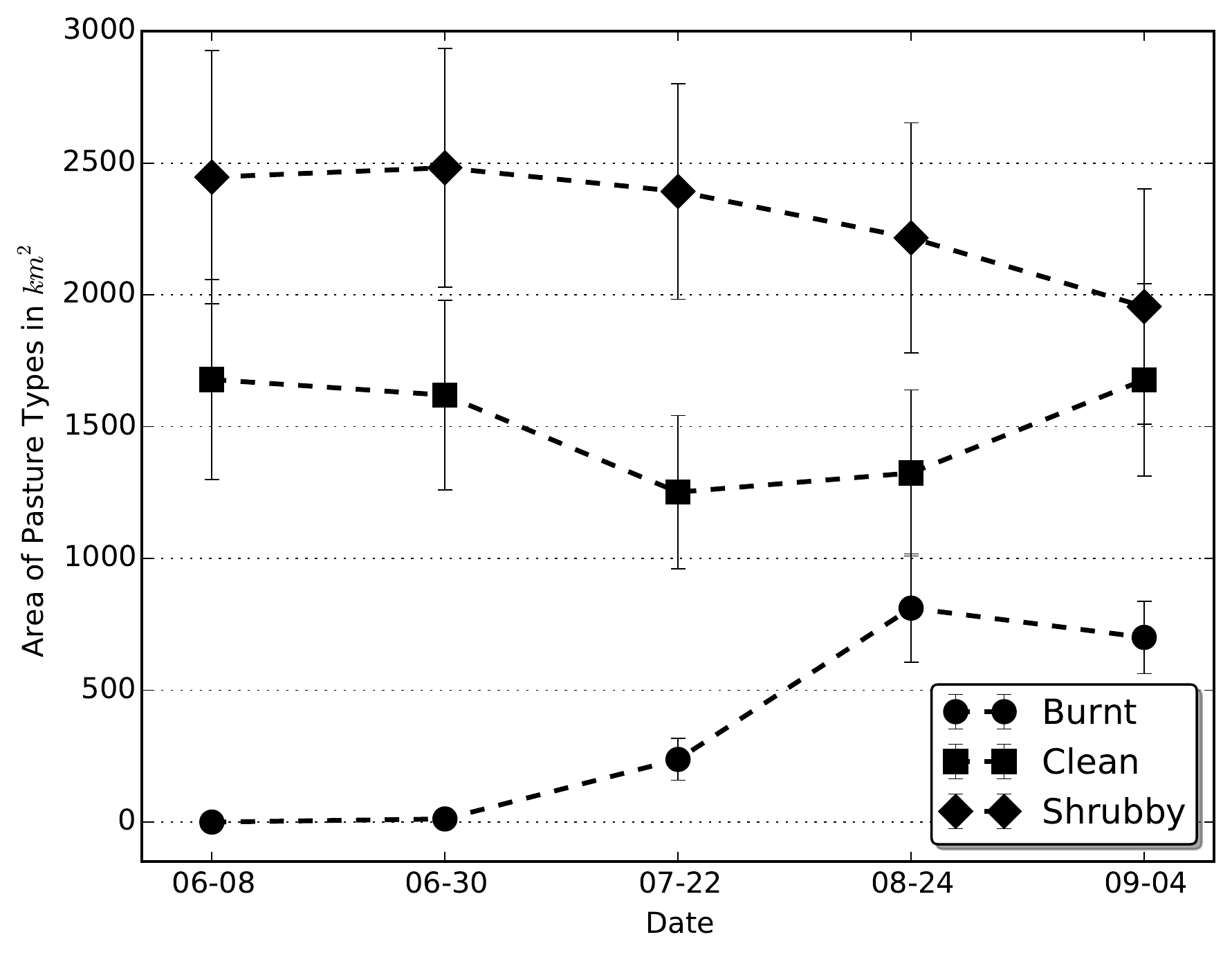}
    \caption{\label{fig:areas}Growth of burnt areas over the 2014 dry season. Error bars are indicative of the $95\%$ confidence interval.}
\end{figure}

\begin{landscape}
	\begin{figure}
		\centering
		\begin{subfigure}[b]{0.25\textwidth}
			\includegraphics[width=\textwidth]{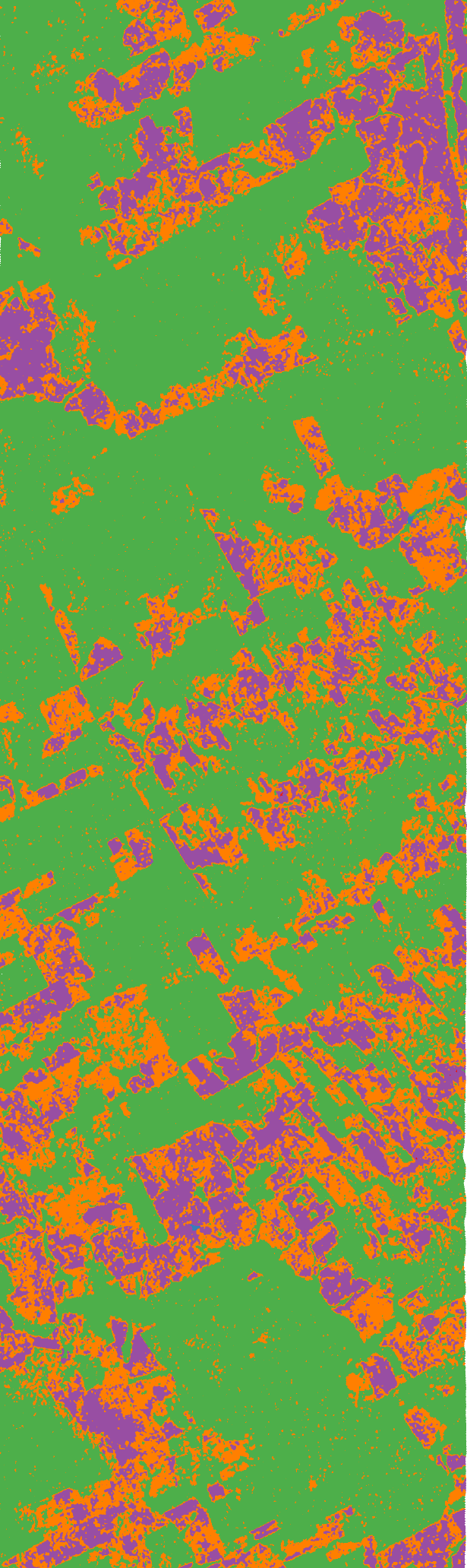}
			\caption{2014-06-08}
			\label{fig:gull2}
		\end{subfigure}
		~ 
		\begin{subfigure}[b]{0.25\textwidth}
			\includegraphics[width=\textwidth]{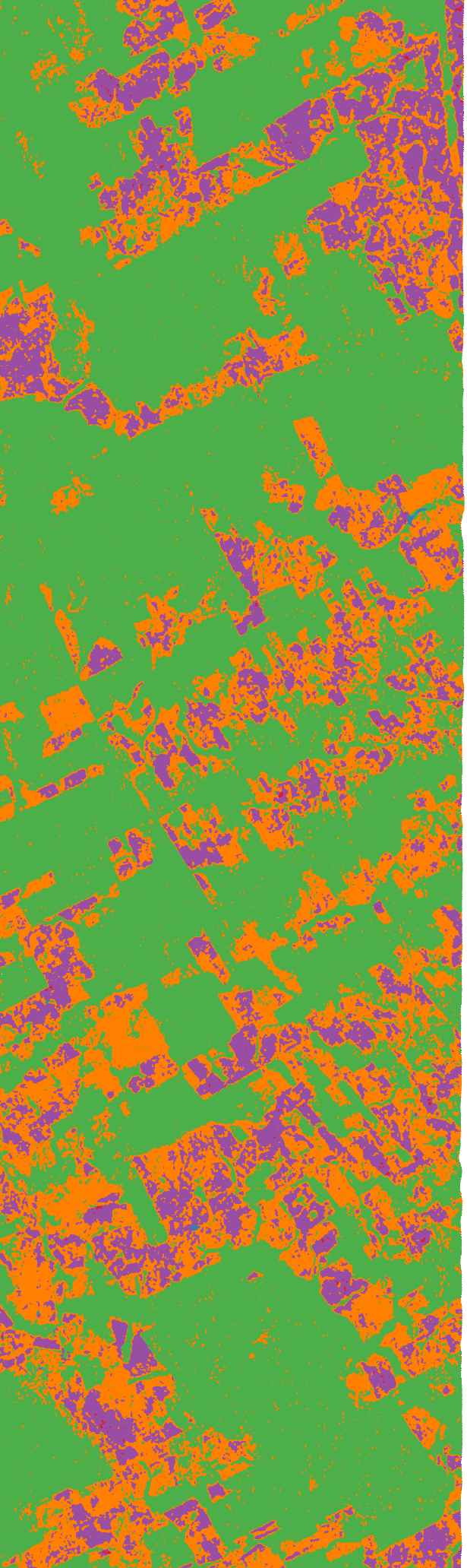}
			\caption{2014-06-30}
			\label{fig:tiger}
		\end{subfigure}
		~ 
		\begin{subfigure}[b]{0.25\textwidth}
			\includegraphics[width=\textwidth]{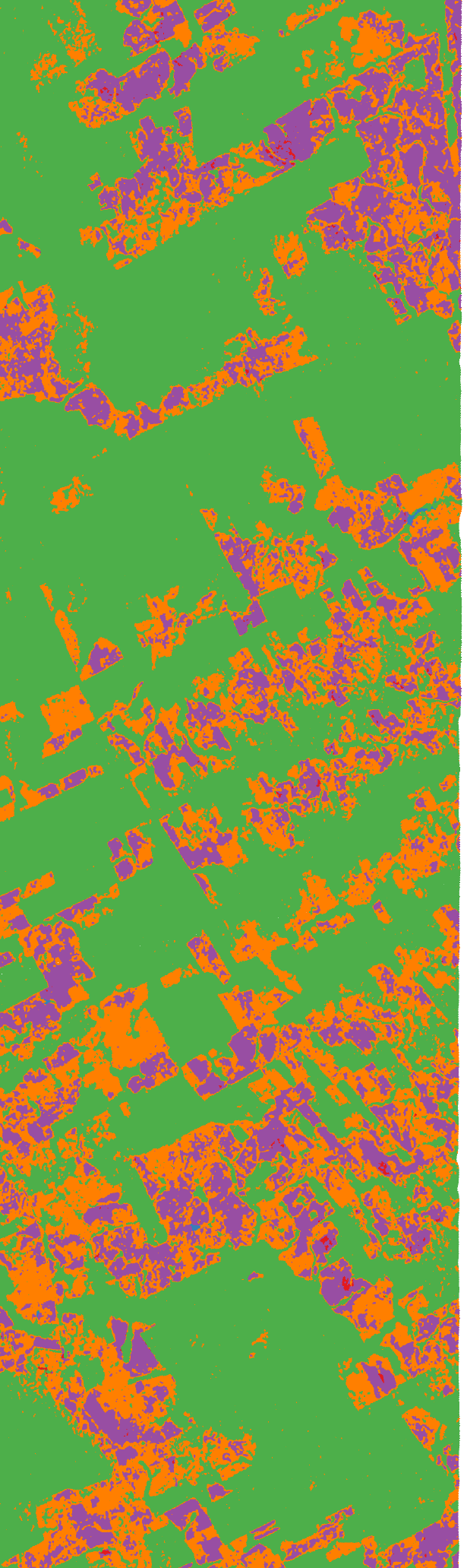}
			\caption{2014-07-22}
			\label{fig:mouse2}
		\end{subfigure}
		~ 
		\begin{subfigure}[b]{0.25\textwidth}
			\includegraphics[width=\textwidth]{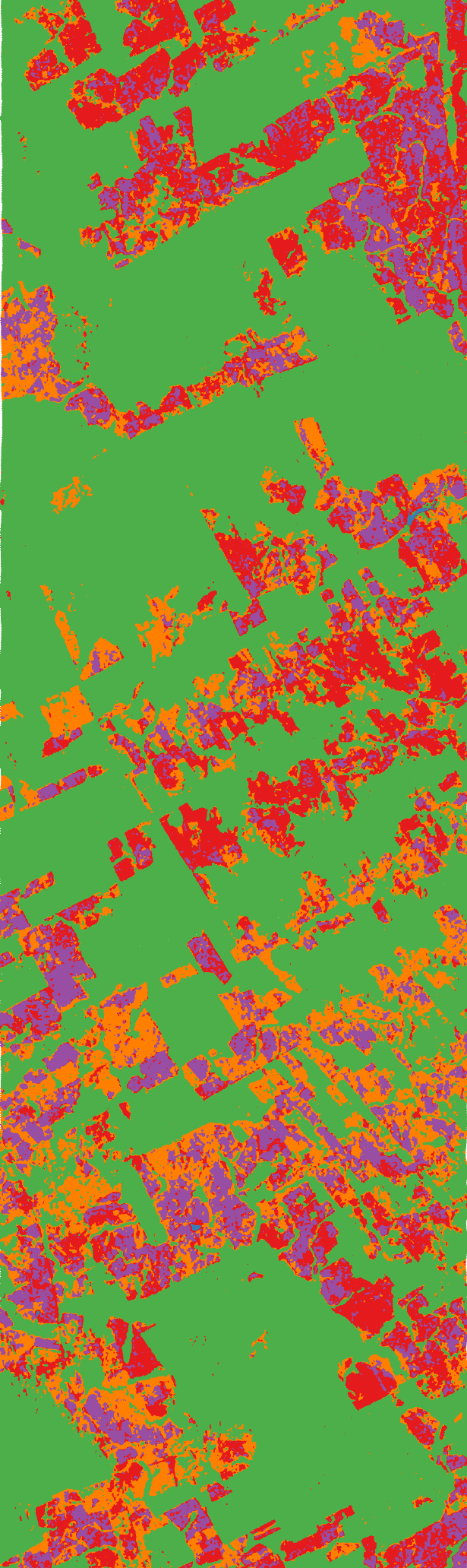}
			\caption{2014-08-24}
			\label{fig:tiger2}
		\end{subfigure}
		~ 
		\begin{subfigure}[b]{0.25\textwidth}
			\includegraphics[width=\textwidth]{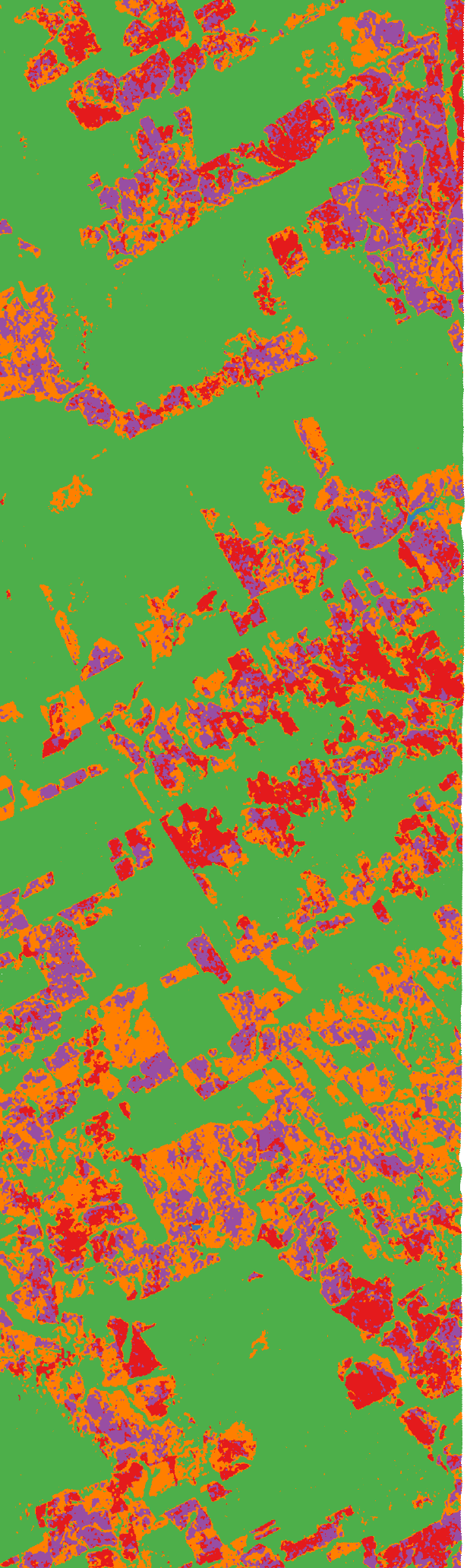}
			\caption{201409-04}
			\label{fig:mouse}
		\end{subfigure}
		\caption{Final classification results using the st-MRF approach.}\label{fig:finalmaps}
	\end{figure}
\end{landscape}

\begin{table}
	\small
	\centering
	\caption{\label{tab:results}Area adjusted overall accuracies for different dates. The shown values are means over 10 iterations.}
	\begin{tabular}{lcccc}
		\toprule
		Acquisition Date & Polarization & IVM & s-MRF & st-MRF \\
		\midrule         
        2014-06-08 & VV-VH & $0.65$ & $0.75$ & $\mathbf{0.77}$ \\
        2014-06-30 & HH-HV & $0.60$ & $0.69$ & $\mathbf{0.79}$ \\
        2014-07-22 & VV-VH & $0.66$ & $0.76$ & $\mathbf{0.78}$ \\
        2014-08-24 & VV-VH & $0.69$ & $0.74$ & $\mathbf{0.76}$ \\
        2014-09-04 & HH-HV & $0.68$ & $0.77$ & $\mathbf{0.78}$ \\
        \bottomrule
	\end{tabular}
\end{table}
\begin{figure}
\centering
	\includegraphics[height=\textheight]{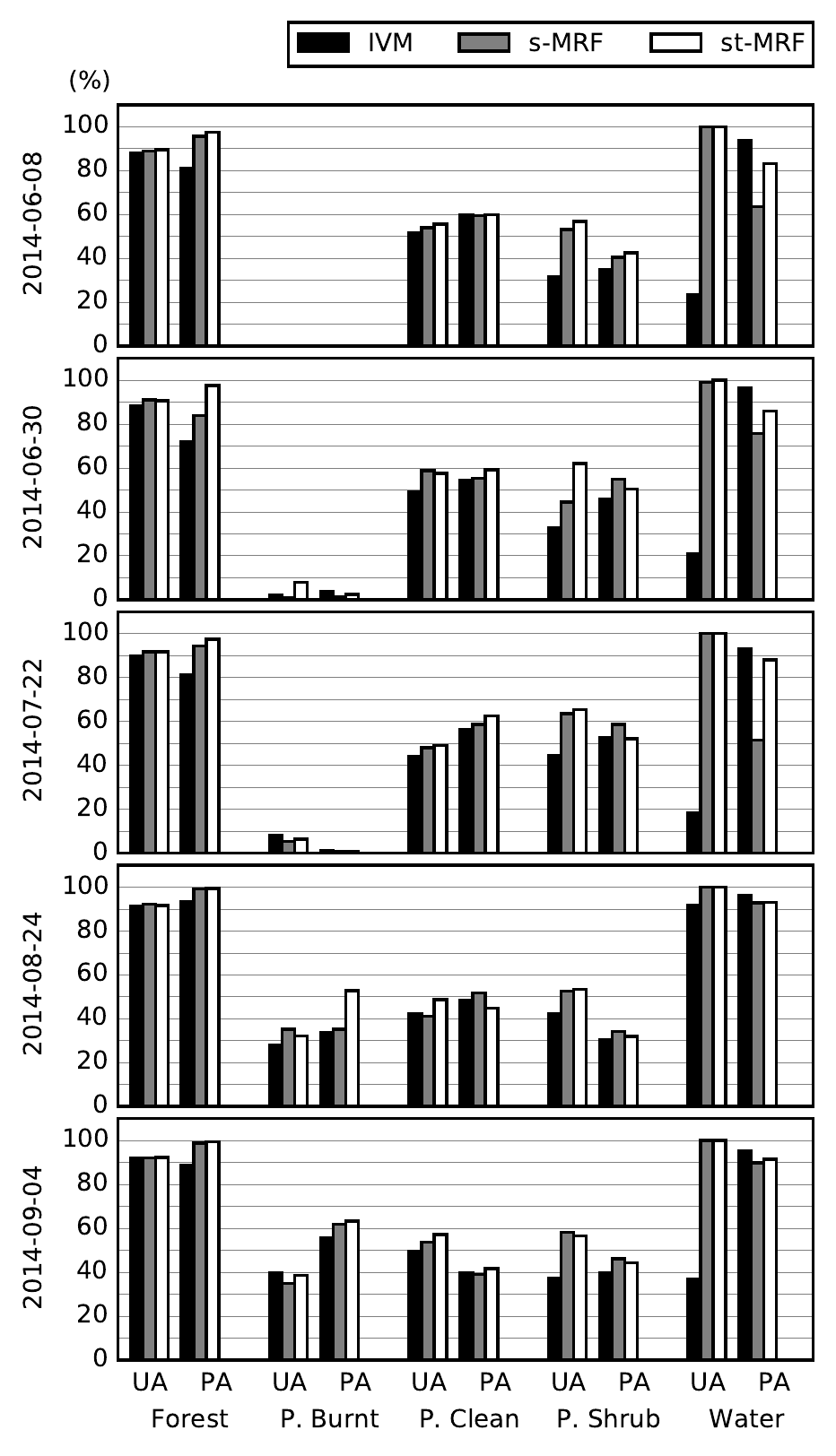}
	\caption{\label{fig:pasanduas}User's and Producer's Accuracy for all the classes at each date.}
\end{figure}
We show that we can benefit from the high repetition rate and high ground resolution of TS-X, and that the proposed framework outperforms common classification approaches in terms of area adjusted mapping accuracy \citep{Olofsson2014}. Table \ref{tab:results} illustrates the average area adjusted OA for the five TS-X scenes, using the three different methods. Irrespective of the acquisition date, the accuracy was significantly improved by the MRF, with the s-MRF consistently outperforming the IVM only results, and st-MRF consistently outperforming s-MRF to a lesser degree. The weakest classification of IVM ($60\%$) and s-MRF ($69\%$) could be clearly improved by up to $19$ and $10$ percentage points compared to the classification results achieved with st-MRF ($79\%$). On average, OA could be improved by $8.6$ percentage points using s-MRF, and $12.2$ percentage points using st-MRF when compared to the IVM classification. As recommended by \citet{Olofsson2014}, we additionally calculated variance measures for results, yet with confidence intervals generally falling well below $1$ percentage point, we will not further address these measurements.\\
Figure \ref{fig:pasanduas} summarizes the average UA's and PA's of the three approaches (IVM, s-MRF, st-MRF). IVM yields the lowest accuracies, while st-MRF generally shows the highest, and most balanced class-specific accuracies. All approaches are especially reliable concerning the classification of forest areas, with st-MRF achieving especially high PA's for this class ($\geq90\%$). For any given approach, the three pasture classes are classified with significantly lower accuracies than forest areas. \texttt{Shrubby pasture} and \texttt{clean pasture} are overall underrepresented, with PA's between $30\%$ and $60\%$ depending on the scene and method. \texttt{Clean pasture} generally yields accuracies of approximately $50\%$, but also classification of this class is particularly problematic concerning the PA of the last scene (around $40\%$). Weak classification results of the different pasture classes are generally caused by confusion within the different pasture types, and reflects findings of comparable studies which utilize X-band SAR data \citep{schlund2013importance}. In general, st-MRF shows higher accuracies, when compared to the classification results achieved by the other approaches, and is capable of mapping \texttt{burnt pasture} starting from 2014-08-24. This is very notable, as a general concern regarding multi-temporal MRF's are its smoothing effects, which could cause the suppression of sporadic events. Due to the low number of \texttt{burnt pasture} areas before the end of July, we are not able to reliably calculate accuracies for \texttt{burnt pasture} areas at every date. Only up to two burnt fields exist for the first two acquisitions, which do not allow for an adequate classification and validation. However, this is also in accordance to the typical land management in the region, insofar slash-and-burn activities usually start later in the season. Nevertheless, the class is kept as st-MRF utilizes any class for the scenes and to have a consistent classification scheme over the entire period. Although some additional \texttt{burnt pasture} areas occur in July (ten areas over the entire study area), the classification accuracy remains very low. Despite consisting of few samples, possibly due to its temporal consistency and very distinct signature, \texttt{water} is mapped especially well. As \texttt{water} encompasses just very few areas over the entire study site, yet yields PAs of $85\%$ and higher for the baseline approach. This weakness appears to get enhanced by the s-MRF approach, which yields a remarkable drop in the PA of \texttt{water} at some dates, while the spatio-temporal MRF appears to ensure its further designation.
This behavior underlines capabilities of st-MRF to not only increase mapping accuracy of temporally sporadic classes, such as \texttt{burnt pasture}, but, remarkably, also proves the value of st-MRF regarding the mapping of classes which are static, yet spatially small scaled. Contrastingly, using mono-temporal MRF such classes tend to get suppressed more frequently. Regarding UA, water is mapped very convincingly with accuracies of over $95\%$ using the MRF approaches, yet the IVM classification shows much less reliable accuracies.
\begin{table*}
\centering
\begin{tabular}{cllll}
& \multicolumn{4}{c}{\includegraphics{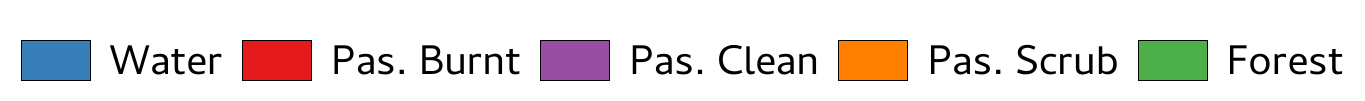}}\\
& IVM & s-MRF & st-MRF & Reference\\
\rotatebox[origin=l]{90}{\makebox[1in]{2014-06-08}}&
\includegraphics[width=0.22\textwidth]{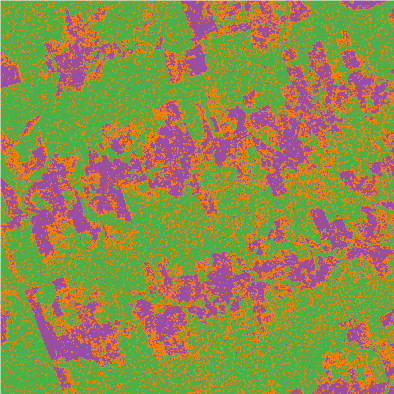}&
\includegraphics[width=0.22\textwidth]{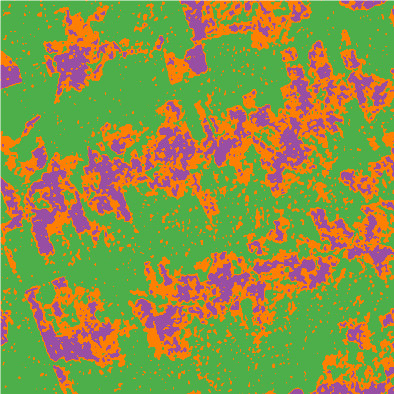}&
\includegraphics[width=0.22\textwidth]{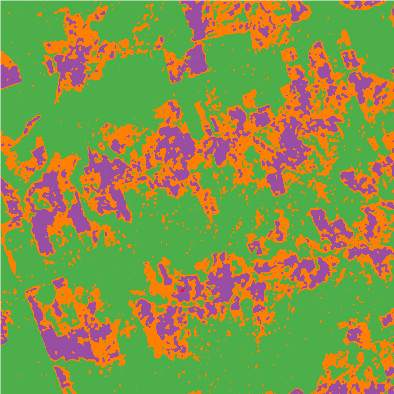}&
\includegraphics[width=0.22\textwidth]{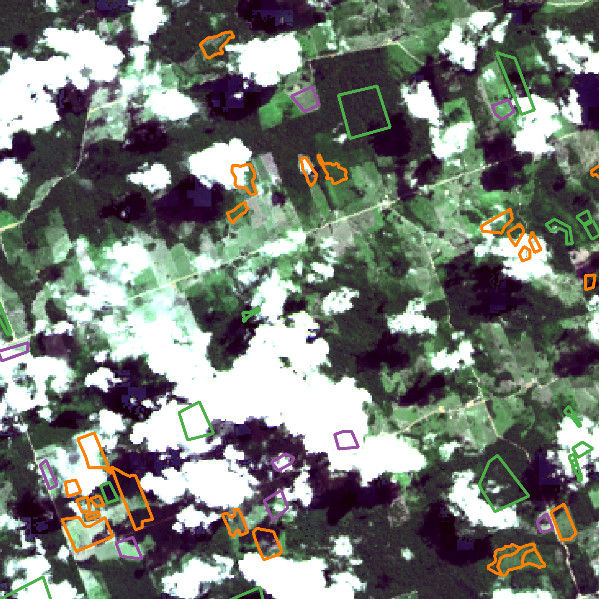}\\
\rotatebox[origin=l]{90}{\makebox[1in]{2014-06-30}}&
\includegraphics[width=0.22\textwidth]{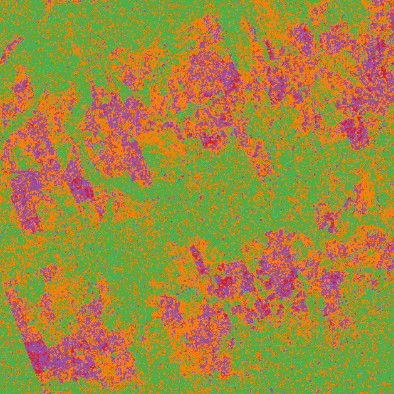}&
\includegraphics[width=0.22\textwidth]{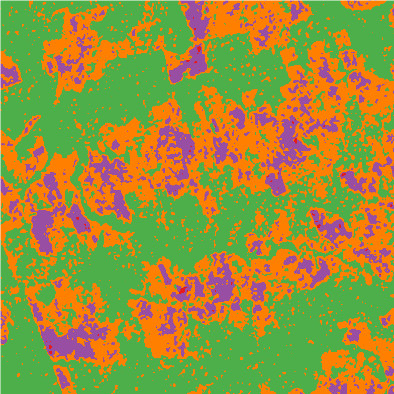}&
\includegraphics[width=0.22\textwidth]{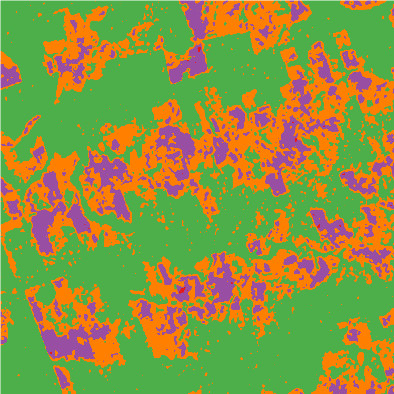}&
\includegraphics[width=0.22\textwidth]{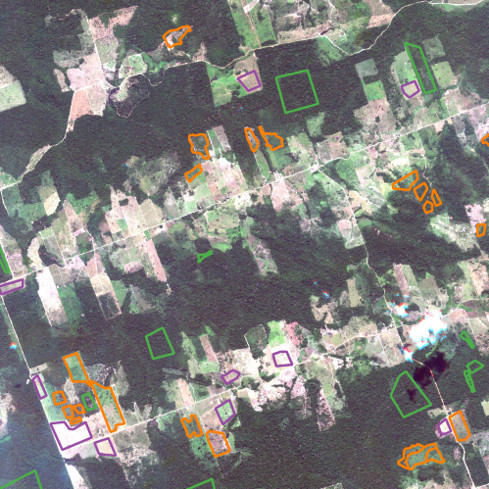}\\\
\rotatebox[origin=l]{90}{\makebox[1in]{2014-07-22}}&
\includegraphics[width=0.22\textwidth]{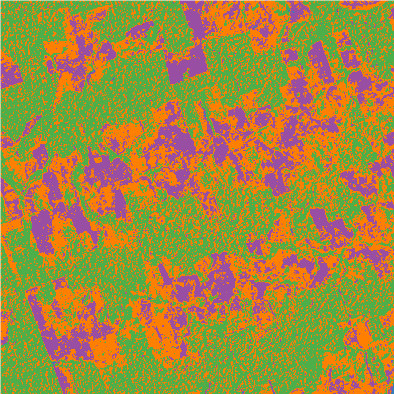}&
\includegraphics[width=0.22\textwidth]{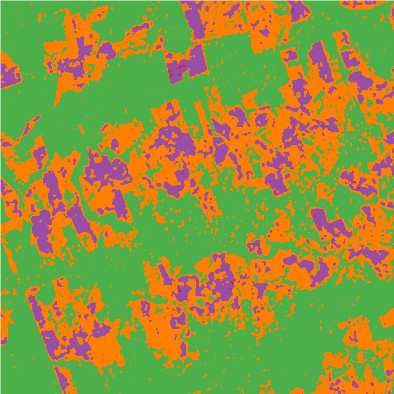}&
\includegraphics[width=0.22\textwidth]{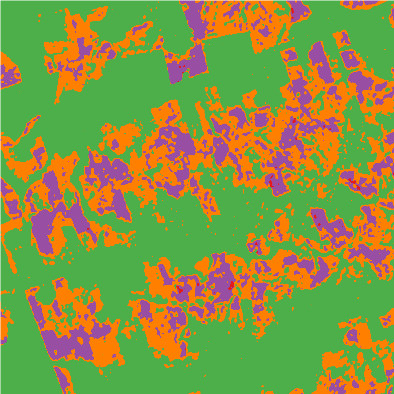}&
\includegraphics[width=0.22\textwidth]{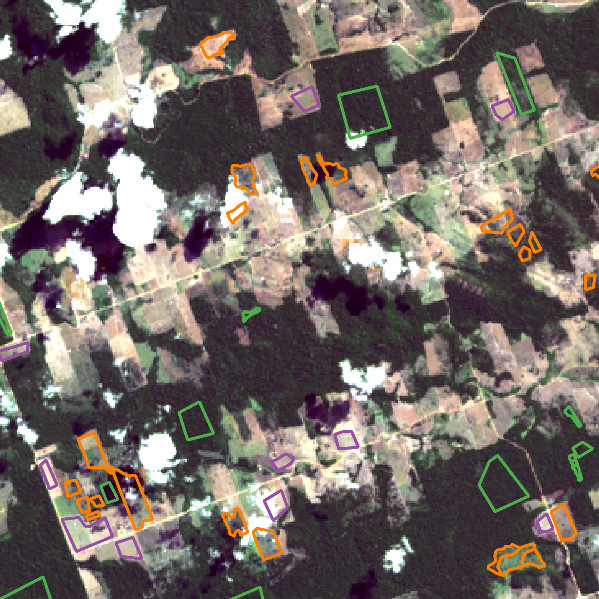}\\\
\rotatebox[origin=l]{90}{\makebox[1in]{2014-08-24}}&
\includegraphics[width=0.22\textwidth]{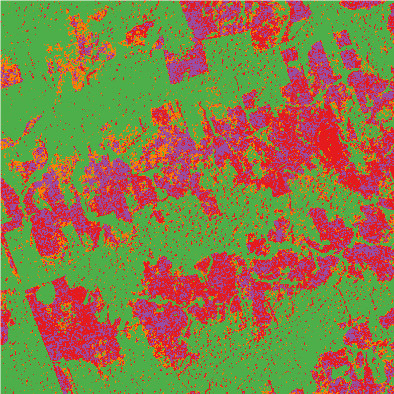}&
\includegraphics[width=0.22\textwidth]{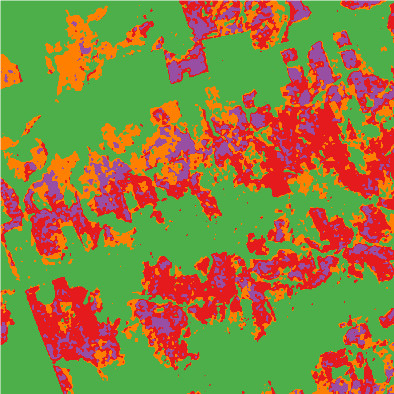}&
\includegraphics[width=0.22\textwidth]{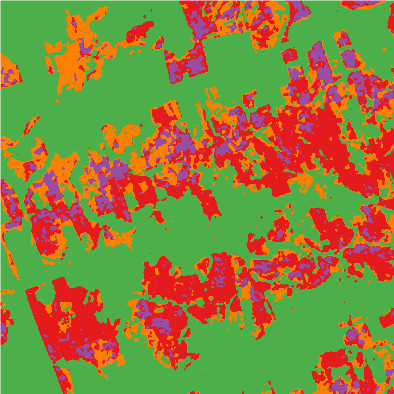}&
\includegraphics[width=0.22\textwidth]{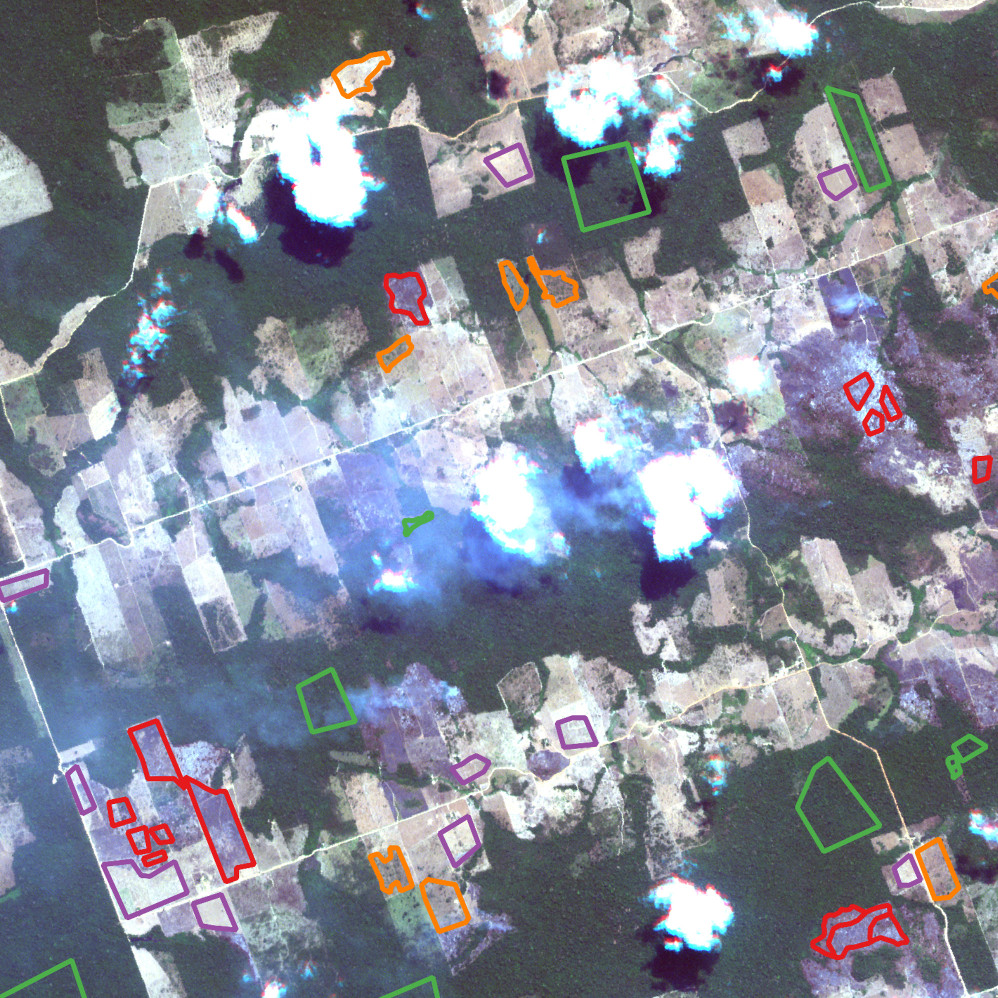}\\\
\rotatebox[origin=l]{90}{\makebox[1in]{2014-09-04}}&
\includegraphics[width=0.22\textwidth]{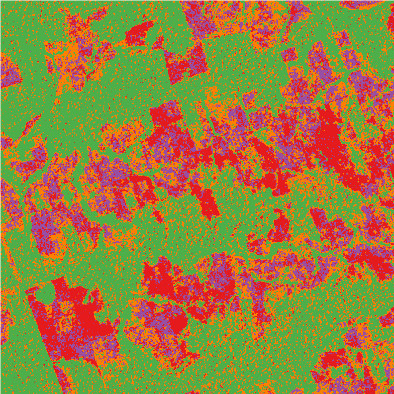}&
\includegraphics[width=0.22\textwidth]{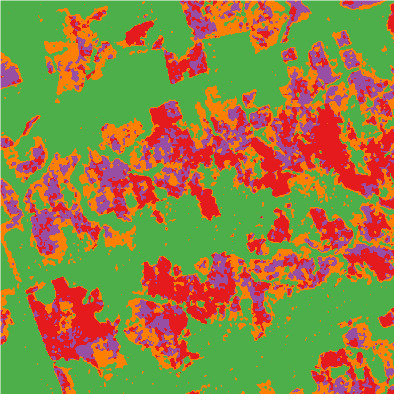}&
\includegraphics[width=0.22\textwidth]{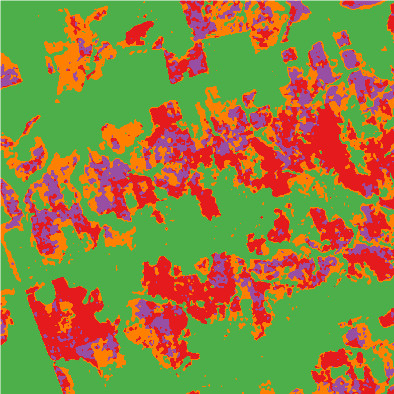}&
\includegraphics[width=0.22\textwidth]{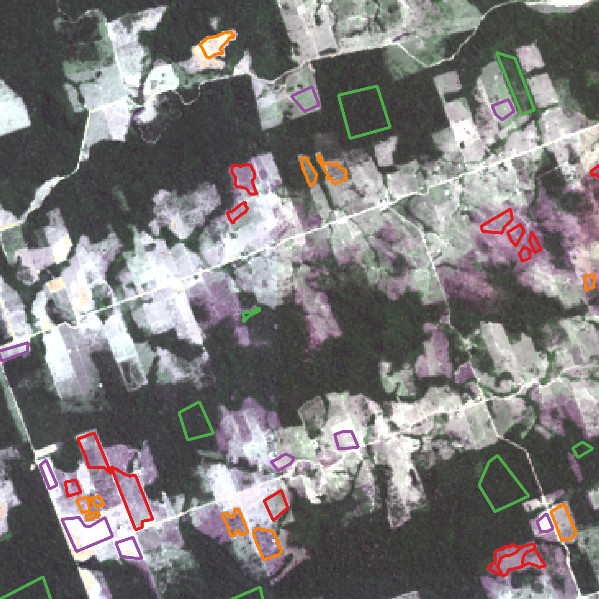}\
\end{tabular}
\captionof{figure}{Comparison of different classifications inside the subsetted area of Figure \ref{fig:tsxsaarea}.\label{fig:maps}}
\end{table*}
\\The visual assessment of the classification maps underlines the positive effect of the MRF-based approaches (Figures \ref{fig:maps} and \ref{fig:finalmaps}). A large number of speckle induced misclassification can be attributed to any of the maps classified using IVM with texture parameters. This effect is suppressed to some extent by s-MRF, yet individual clusters of misclassification can still be located. While not entirely homogeneous, st-MRF suppresses the noise considerably, yet maintaining the general spatial patterns of LULC. Fine spatial structures appear to not get suppressed by the st-MRF, despite conservative IVM estimates.\\Using land cover maps derived from st-MRF, Figure \ref{fig:areas} illustrates that clear trends can still be derived using the proposed data and methods. The figure illustrates a high percentage of shrubby pasture land early in the dry season. Over the course of the dry season, this amount is continuously shrinking, while the burning of pasture starts growing exponentially at the end of July. At the end of the dry season, the area of clean pasture land is comparable to that of shrubby pasture.
\section{Discussion}\label{disc}
\begin{figure}
	\includegraphics[width=\textwidth]{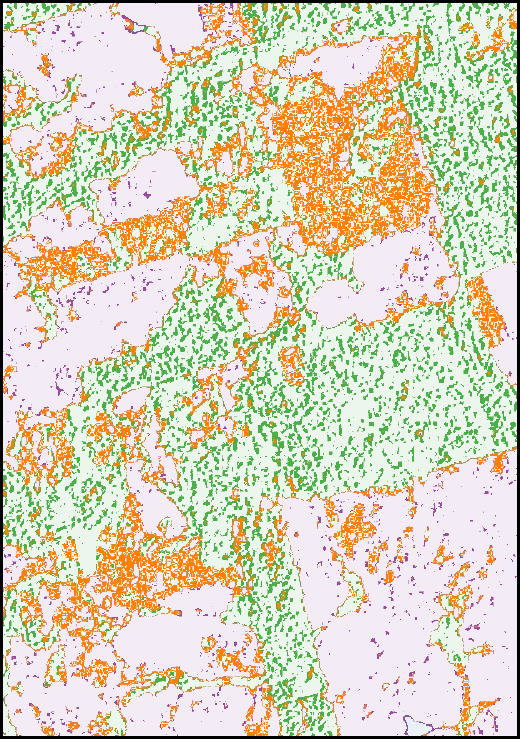}
    \caption{\label{fig:diffim}Difference map between 2014-06-30 classifications of st-MRF and IVM. Light colors indicate agreement between the two maps. Dark colors indicate class ambiguities, while the class of the final st-MRF classification is presented.}
\end{figure}
The main objective of this study, i.e., the adaptation of recent methods for the mapping of dynamic LULC in a tropical setting, is shown to be generally positive in our study. The proposed approach of using spatial-temporal MRF with expert knowledge is generally able to capture short term LULC dynamics, which are challenging to map using standard classification techniques. Although our validation confirms limitations of short-wavelength SAR data (e.g., when differentiating different pasture types, especially mono-temporally), the proposed approach enables the generation of a meaningful time series of homogeneous LULC maps using only SAR data. Consequently, the reliable prompt mapping of LULC change can be achieved independent of cloud cover and atmospheric inference.\\
The results show that the use of the proposed approach outperforms standard IVM classifications utilizing texture parameters only, as well as common spatial MRF, in terms of classification accuracy. Visual inspection of burnt pasture areas of early dates shows bright and overall heterogeneous backscatter within the class, and similarity to the other pasture classes at TS-X images, while Landsat and RapidEye images unambiguously indicate burnt pasture. Possible reasons for this could be organic debris or wet conditions. Contrary to that, many burnt areas of subsequent scenes, after the occurrence of large scale burning, can be identified more clearly at X-band as areas of low backscatter.\\Regarding X-band data the potential transfer of the approach to the wet season, which is characterized by higher saturation of backscatter intensity, is another challenge. While separation of pasture types already appears difficult in the dry season, the integration of temporal context via the MRF might allow for a reliable separation of pasture and forest areas over the wet season. Additional testing showed furthermore that utilization of temporal trajectories alone, despite generally not as effective as utilization of the spatial context (s-MRF), can be used to significantly elevate all accuracies above $70\%$. In particular the weak classification of 2014-06-30 could benefit from this approach as the variance of classification outcomes is reduced between different scenes.\\
Our findings are in accordance with the results of other recent studies, which were able to improve the classification accuracy via implementation of multi-temporal MRF \citep{wehmann2015spatial,liu2008using}. While \citet{wehmann2015spatial} use regionally optimized transition matrices, and a state of the art integrated kernel based on \citet{moser2013combining} to achieve high classification accuracies over long time periods, the proposed method aims on the detection of short term land cover change in SAR imagery, and utilizes LBP for inference as well as IVM for classification.
The visual assessment of the classification results confirms the positive effect of the MRF on the classification accuracy. Although the maps provided by the conventional IVM classification show general land use patterns, the results are affected by typical SAR-inherent noise. Even homogeneous areas appear very noisy, despite texture parameters that were included in the classification procedure.
Boundaries between individual land cover and land use classes may appear blurred and are hard to identify. This drawback is significantly reduced by both MRF-based methods. As LBP tries to minimize the transition energy by homogenizing adjacent pixels, areas become overall more concentrated and edges along different LULC classes can be more clearly identified. Benefits can also be attributed to the classification of the interior of areas, as the application of MRF suppresses outliers. Thus, the results confirm the edge preserving capabilities of MRF, even for challenging spatial class transitions (e.g. \texttt{forest} to \texttt{shrubby pasture}).\\With regard to class-specific accuracies, the spatio-temporal MRF offers preferable results over just IVM and the spatial-only approach. Figure \ref{fig:diffim} illustrates the differences between a 2014-06-30 classification of the IVM and the st-MRF approaches, underlining the potentials of solving confusion between \texttt{forest} and \texttt{shrubby pasture}. It is colorized to highlight disagreements of the classifications, with pale colors signifying consenting classifications, and opaque colors indicating classes as assigned by st-MRF (for legend see Figure \ref{fig:maps}). It is especially obvious that with increasing vegetation density confusion also rises, and that \texttt{clean pasture} is classified congruently in both approaches. While the classification of \texttt{clean pasture} and \texttt{shrubby pasture} remains challenging, TS-X data constitutes an adequate data source for \texttt{forest} / \texttt{non-forest} mapping. The PA and UA for forest are higher compared to the accuracies achieved for the other classes, and are in accordance with the accuracies of comparable studies \citep{schlund2013importance,garcia2011analise}. While we did not perform specific analysis on the differences of HH-HV and VV-VH polarized data sets, Table \ref{tab:results} shows the OA of the internal HH-HV polarized scenes (06-30) to especially benefit from the multi-temporal integration, and also its neighboring scenes to benefit disproportionately. We can thus assume synergetic effects to be transferable through a multi-temporal MRF, yielding a promising outlook for the multi-sensoral integration of various data sources.\\
Regarding the low requirements concerning parameterization and the implementation through moving windows, we consider the introduced method to be transferable to other study regions. Adaptation of the transition matrices allows the method to be fitted to more static environments, or also to address multi-annual time series data. Despite the ambitious goals of this study, i.e. to perform land cover mapping in a densely vegetated and dynamic tropical study region using TS-X data, and some documented limitations concerning the separability of different pasture types, we were able to achieve improvements over standard classifications. As the method incorporates adjacency information, potential shortcomings exist when the ground resolution is coarse relative to the mapped land cover. In this case, fragmented structures might get suppressed. Further adjustment would also be required if assumptions on land cover trajectories are variant in a multi-temporal setup. For example, two scenes from the dry season carry a different transition probability with regard to burning than two scenes from the wet season. While this can be easily solved through different transition models, within this study we just included slight modifications of transition matrices to account for the different intervals between TS-X acquisitions.
\section{Conclusion}\label{conc}
The results show clearly that the integration of spatial-temporal MRFs are advantageous to the baseline classification approach and spatial MRF methods. Especially the classification of forest areas yields very high accuracies. We were able to successfully implement an LBP optimization for the regularization of high resolution, multi-temporal TS-X images of a tropical context. We were furthermore able to give adequate estimates on the spatio-temporal pattern of land use dynamics such as burned pastures. Importantly, the suggested approach is able to handle process of small spatio-temporal scale, and despite its smoothing effects does not suppress fine structures. Separation of different types of pasture (pasto sujo and pasto limpo) remains a challenging task at the short wavelength. Classification of \texttt{burnt pasture} early in the season highlights limitations of the MRF-based model, which arise when the underlying classification accuracy is already limited. While the approach is well suited to regularize small classification errors using contextual information, it is not able to sufficiently address misclassifications in complex, transitional environments with weak classification accuracies. The sometimes relatively low class accuracies are not necessarily a limitation of the proposed method, but rather caused by the short-wave TS-X data as well as class-specific characteristics. \\
Particularly for study sites which are characterized by land use patterns of high spatial-temporal variability, the proposed approach (i.e., using spatial-temporal MRF with expert knowledge) appears feasible. Using expert knowledge on land cover trajectories, we could positively influence model performance and bypass computationally demanding techniques for the estimation of MRF parameters.\\When derived from multiple classifications, change maps are generally strongly affected by weak initial classifications.
\\The proposed method is formalized to be transferable to large, possibly multi-sensoral, image stacks.
For future studies our aim is to integrate the regularization of short-term, intra-annual dynamics with long-term dynamics such as deforestation and agricultural trends, using multi-sensoral imagery.\\%
\section*{Acknowledgments}
The study was carried out as part of the SenseCarbon research project funded by DLR/BMWi (50 EE 1255). SenseCarbon is part of GFOI-R\&D (BRA-2). We thank the countless developers behind the freely accessible software utilized in this study (i.e. Python, GDAL, S1TBX, and Q-GIS). For their supporting efforts we direct further thanks to Niklas Potthoff, Paul Wagner, Rolf Rissiek, Bernd Melchers, Merry Crowson, and Christian Lamparter, as well as the anonymous reviewers.
\section*{References}
\bibliographystyle{elsarticle-harv} 
\bibliography{database.bib}

\end{document}